%% file: main.tex
\documentclass[runningheads]{llncs}

\usepackage{eccv}

\usepackage{eccvabbrv}

\usepackage{graphicx}
\usepackage{booktabs}

\usepackage[accsupp]{axessibility}  

\input{preamble}

\usepackage{hyperref}

\begin{document}

\title{EGM: Efficient Visual Grounding \\ Language Models}

\titlerunning{EGM}

\author{\parbox{\textwidth}{\centering
  Guanqi Zhan$^{1,3*}$, Changye Li$^{4*}$,
  Zhijian Liu$^{1}$, Yao Lu$^{1}$, \\ Yi Wu$^{4}$, Song Han$^{1,2}$, Ligeng Zhu$^{1}$\\[4pt]
    $^1$NVIDIA\quad\quad
    $^2$MIT \quad\quad
    $^3$University of Oxford\quad\quad
    $^4$Tsinghua University\\[4pt]
    \normalsize{Github: \url{http://github.com/NVLabs/EGM}} \\
    \normalsize{Website: \url{https://nvlabs.github.io/EGM}} \\
    \normalsize{Models: \url{https://huggingface.co/collections/nvidia/nvidia-egm}}
}}
\authorrunning{G. Zhan, C. Li et al.}
\institute{}

\maketitle

\input{sec/0_abstract}    
\input{sec/1_intro}

\input{sec/2_related}
\input{sec/3_problem_analysis}

\input{sec/4_method}

\input{sec/5_experiment}

\input{sec/6_conclusion}

\section*{Acknowledgements} 
We thank Prof. Andrew Zisserman, An-Chieh Cheng, Haian Huang, Sifei Liu, Baifeng Shi, Hongxu Yin, Hanrong Ye, Wenwei Zhang for their suggestions, help and support for the project.

\bibliographystyle{splncs04}
\bibliography{main}

\input{sec/7_supple}

\end{document}

%% file: preamble.tex
\usepackage[table]{xcolor} 
\definecolor{yellowgreen}{rgb}{0.4,0.8,0.1}
\usepackage{multirow}
\usepackage[normalem]{ulem}
\newcommand*\failreason{`COMPLEX-PROMPT'\xspace}
\newcommand*\method{EGM\xspace}

%% file: sec/0_abstract.tex
\begin{abstract}

Visual grounding is an essential capability of Visual Language Models (VLMs) to understand the real physical world. Previous state-of-the-art grounding visual language models usually have large model sizes, making them heavy for deployment and slow for inference. However, we notice that the sizes of visual encoders are nearly the same for small and large VLMs and the major difference is the sizes of the language models. Small VLMs fall behind larger VLMs in grounding because of the difference in language understanding capability rather than visual information handling. To mitigate the gap, we introduce \emph{`\textbf{E}fficient visual \textbf{G}rounding language \textbf{M}odels' (\method)}: generate many mid-quality tokens (from small models) to match the performance of large VLMs with few high-quality but expensive tokens. This method is deployment-friendly, and yields better end-to-end latency: On the RefCOCO benchmark, our \textbf{\method-Qwen3-VL-8B} demonstrates \textbf{91.4 IoU} with an average of 737ms (\textbf{5.9$\times$ faster}) latency while \textbf{Qwen3-VL-235B} demands 4,320ms to reach \textbf{90.5 IoU}. To validate our approach's generality, we further set up a new amodal grounding setting that requires the model to predict both the visible and occluded parts of the objects. Experiments show our method consistently improves both vanilla and amodal grounding capabilities of small models to match or outperform larger models, thereby improving efficiency for visual grounding. 

\end{abstract}

%% file: sec/1_intro.tex
\section{Introduction}
\label{sec:intro}

\begin{figure*}
    \centering
    \includegraphics[width=1.0\linewidth]{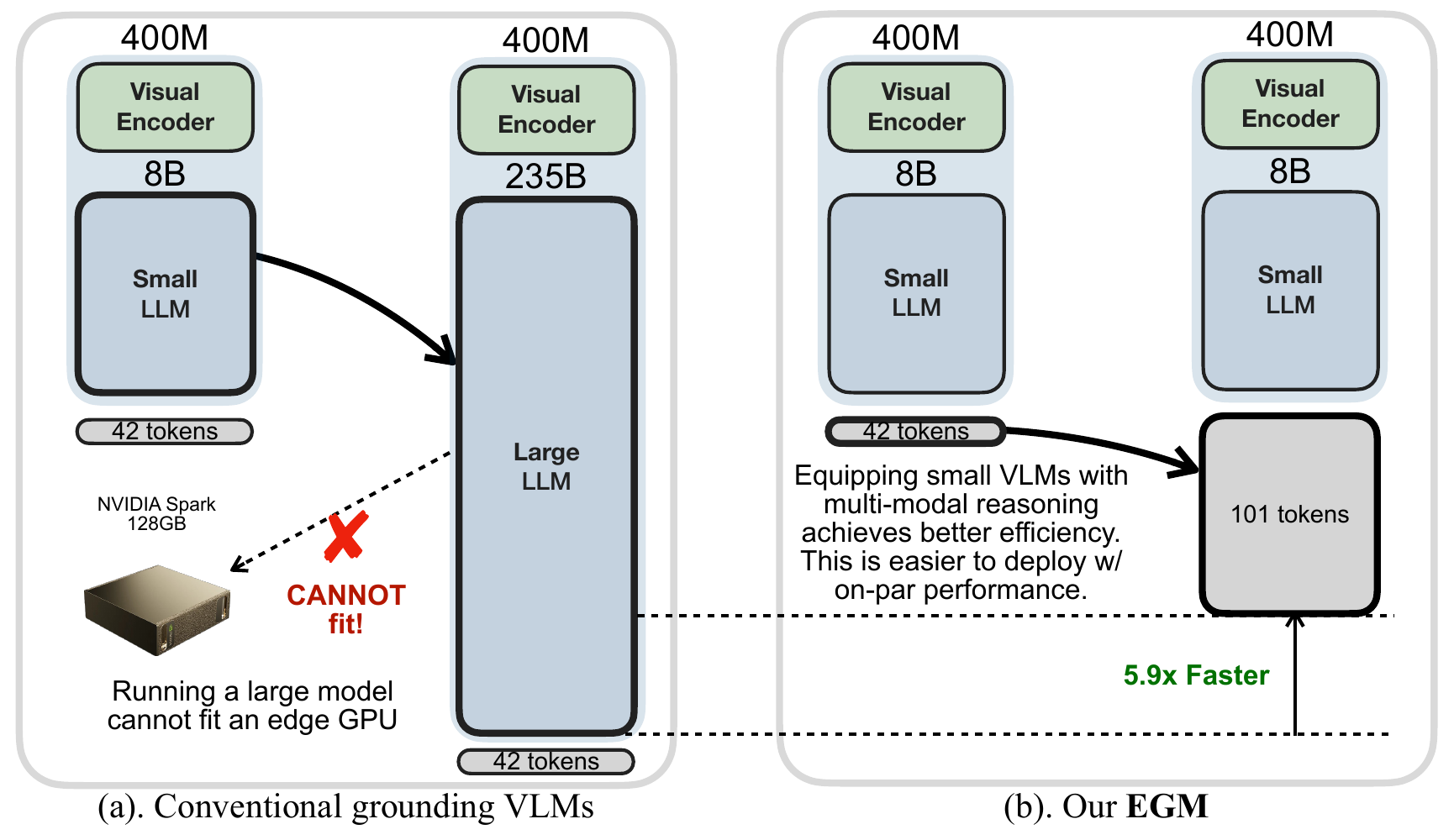}
    \caption{\textbf{Overview of Efficient Visual Grounding Language Models.} \textit{Left.} Existing state-of-the-art grounding VLMs usually have large model sizes. \textit{Right.} Our \method enhances text understanding capabilities of small VLMs by equipping them with multi-modal reasoning capability, achieving better efficiency.}
    \label{fig:main_teaser}
    \vspace{-8pt}
\end{figure*}

Visual grounding is an essential capability of Visual Language Models (VLMs) to understand the physical world, especially for autonomous driving or robotics systems, where it is important to locate the target objects accurately according to the natural language instructions. Previous state-of-the-art visual grounding VLMs usually have large model sizes to achieve the SOTA performance, \emph{e.g.,} InternVL-3.5~\cite{wang2025internvl35} and Qwen3-VL~\cite{bai2025qwen25}. This brings challenges to deployment and is unfriendly to edge systems (e.g., Jetson Thor, Spark) while their small versions suffer from inferior performance.

A natural question then arises here: \textbf{What makes small VLMs fall behind bigger ones for visual grounding?} 
Our investigation begins with the scaling trends of the Qwen family~\cite{bai2025qwen25, alibaba2025qwen3vl}.
The visual encoder is the same across different sized VLMs, and the main upgrade is that larger models are built with a bigger Large Language Model (LLM). We hypothesize that \uline{the gap between different sized VLMs mainly comes from their text understanding ability (LLM) rather than visual information handling (ViT)} as shown in Figure~\ref{fig:main_teaser}. 
We thoroughly verify it by visualizing the failure cases of a small VLM and find that a prominent failure pattern is ``prompts too complicated'', where the text prompt is semantically complicated and there are multiple similar candidates in the image (Figure~\ref{fig:failure_case_analysis}). This failure pattern is gradually improved as the model size increases.

\begin{sloppypar}
This insight suggests a pathway towards overcoming the issue: 
extending the test-time compute to enhance the text understanding ability and mitigate the performance gap with large models.
More specifically, we encourage the VLM to generate the detailed reasoning process before the yielding the grounding results. 
To achieve this, as in Figure~\ref{fig:method}, we use a proprietary VLM to generate detailed reasoning paths toward the target box, creating rich Supervised Fine-tuning (SFT) data for training. After the model learns the reasoning process, it is further trained with Reinforcement Learning (RL) to boost grounding capability.
We find that this paradigm not only boosts the grounding performance of small VLMs but also brings better efficiency and deployment capability: many mid-quality tokens are actually cheaper than a few expensive ones (Figure~\ref{fig:efficiency_tradeoff}).
To further validate the effectiveness of the paradigm, we introduce a new grounding setting -- amodal grounding. This task requires the model to predict both the visible and occluded parts of an object, and we apply our method to this new challenge and also observe solid improvements.
\end{sloppypar}

We conduct experiments on state-of-the-art open-source visual language models including QwenVL and InternVL in Section~\ref{sec:experiments}. 
Results show our method can consistently improve the visual grounding capability of small models and significantly mitigate the gap between small models and bigger models.
We name our models \emph{`\textbf{E}fficient visual \textbf{G}rounding language \textbf{M}odels'(\method)}.

In summary, we make the following contributions:  
\begin{itemize}
    \item We identify the complicated prompt as the main reason why small VLMs lag behind bigger ones. To address this, we introduce \method, which equips small VLMs with multi-modal reasoning abilities through extended test-time computation to enhance their visual grounding capabilities.
    \item We curated SFT and RL data that enables reasoning grounding skills for small VLMs. We also set up amodal grounding with fresh data to further validate our method's generality, challenging the models to predict both visible and occluded parts. 
    \item Our experiments show that \method consistently boosts the grounding performance of small VLMs across sizes and model families, in both vanilla and amodal grounding. For example, on the RefCOCO~\cite{kazemzadeh2014referitgame, mao2016generation} benchmark, our \textbf{8B model} achieves a \textbf{91.4 IoU}, beating the \textbf{235B model's 90.5 IoU} while running \textbf{5.9$\times$ faste}r.
\end{itemize}

%% file: sec/2_related.tex
\section{Related Work}
\label{sec:related_work}

\noindent \textbf{Visual-Language Models (VLMs)} have advanced rapidly recently. Proprietary ones such as GPT~\cite{achiam2023gpt,hurst2024gpt}, Gemini~\cite{team2023gemini,team2024gemini,comanici2025gemini}, Claude~\cite{anthropic2025claude}, and Grok~\cite{xai2025grok4} have greatly improved performance on specific tasks and capability of multiple modalities. On the other hand, open-source VLMs have also developed continuously, closing the gap between open-source and proprietary models. Representative examples include QwenVL~\cite{wang2024qwen2,bai2025qwen25}, InternVL~\cite{chen2024internvl,zhu2025internvl3,wang2025internvl35}, LLaVA~\cite{li2024llava}, Llama~\cite{touvron2023llama,touvron2023llama2,dubey2024llama3}, VILA~\cite{lin2024vila,liu2025nvila} and Molmo~\cite{deitke2024molmo}. The main architecture of these models is LLaVA-like, \emph{i.e.,} the image  is firstly fed into a ViT encoder, and then projected to the LLM via an MLP. The application of VLMs has extended far beyond standard visual question answering. They are now increasingly adopted in complex real-world scenarios, encompassing autonomous driving, robotic manipulation, medical image analysis, and automated web-browsing agents.

\vspace{2pt}
\noindent \textbf{Grounding VLMs.} Visual grounding lets VLMs predict a bounding box for a target object given a text prompt. This is a fundamental capability of VLMs to understand the visual world. Leading open-source models, such as InternVL~\cite{wang2025internvl35} and QwenVL~\cite{bai2025qwen25}, perform best on the RefCOCO~\cite{kazemzadeh2014referitgame,mao2016generation} benchmark, even surpassing proprietary models such as GPT~\cite{hurst2024gpt} and Gemini~\cite{comanici2025gemini}.

Beyond the standard grounding task, \emph{amodal grounding}, predicting both the visible and hidden parts of the target described in the prompt, is also gaining increasing attention. This `amodal' ability reflects how humans perceive objects as complete, even when occluded~\cite{briscoe2011mental,kaup2024modal}. Although amodal completion is gaining attention in computer vision~\cite{li2016amodal,zhu2017semantic,zhan2020self,li20222d,li2023muva,li2023gin,zhan2024amodal,xu2024amodal,chen2025using,liu2025towards,wu2025amodal3r,li2025amodal,lu2025taco}, VLMs have not mastered this skill, even for the most advanced proprietary models. 
In this paper, we are the first to study \emph{amodal grounding} in VLMs, aiming to boost their amodal grounding ability as well as the efficiency.

\vspace{2pt}
\noindent \textbf{Reasoning.} 
Reasoning has become a critical capability of modern LLMs, supporting key applications such as mathematical problem solving and code generation. \cite{guo2025deepseek} introduced an effective reasoning-training pipeline that combined a SFT cold start with Group Relative Policy Optimization (GRPO)~\cite{shao2024deepseekmathpushinglimitsmathematical}, showing that structured reasoning traces together with reward optimization can substantially improve a model's reasoning ability. Following this line of work, a series of GRPO-style algorithms have been proposed. \cite{yu2025dapoopensourcellmreinforcement} developed Dynamic Sampling Policy Optimization (DAPO), which used a token-level mean loss to enable importance sampling and advantage estimation at the token granularity. \cite{tan2025gtpogrpostokensequencelevel} further introduced Group Token Policy Optimization (GTPO), which incorporated entropy regularization into the GRPO framework to encourage more diverse model responses.

%% file: sec/3_problem_analysis.tex
\begin{figure*}[t]
	\centering
\includegraphics[width=1.0\linewidth]{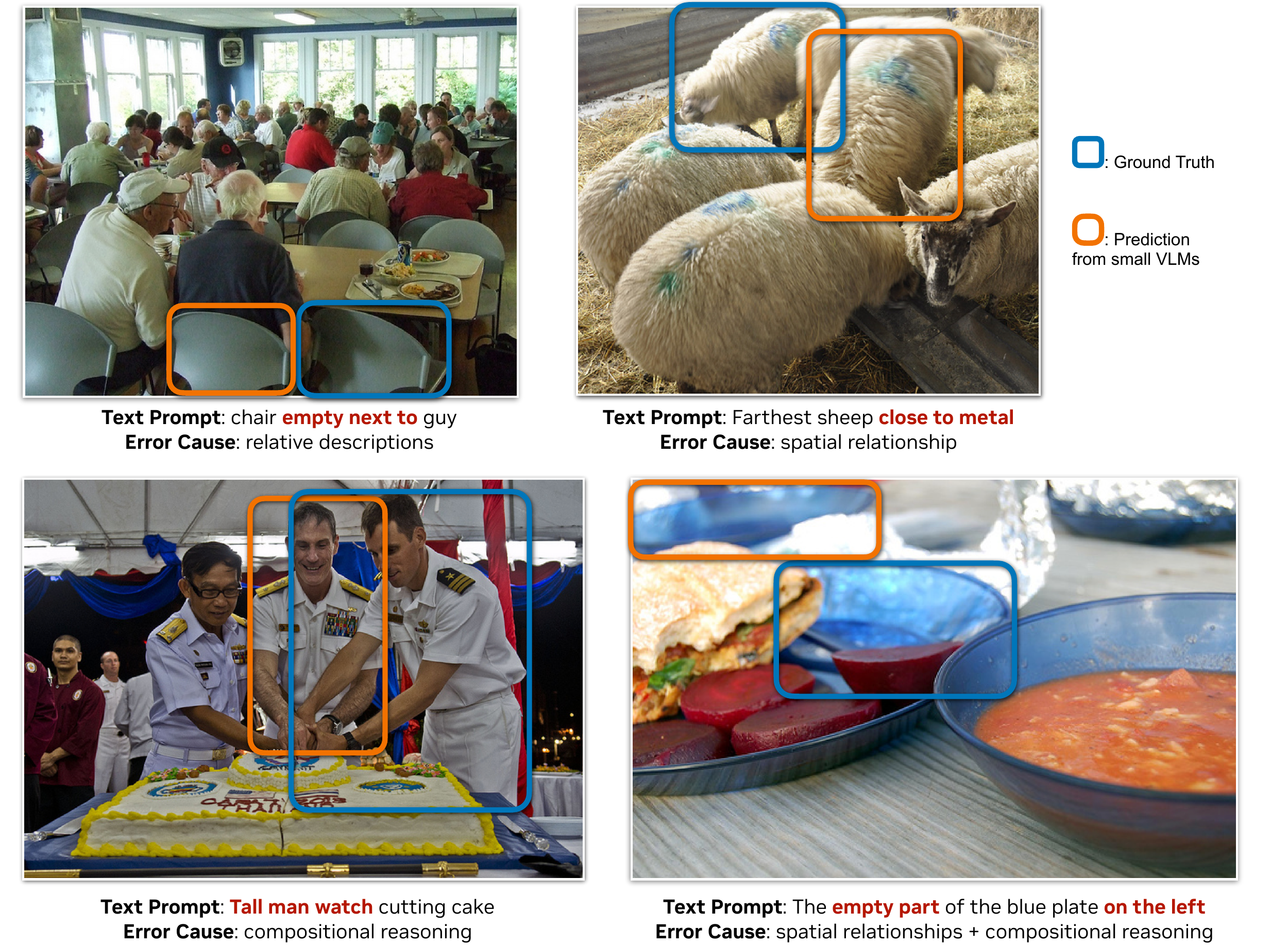}
\caption{
\textbf{Failure cases of small VLMs.} We find small VLMs, \emph{e.g.,} InternVL-3-8B, tend to fail when the text prompt is \textit{semantically complicated} and there are \textit{multiple candidates} in the image that can confuse the model. We term this failure pattern \failreason and label the ground truth bbox in blue, and the 8B model prediction in orange in examples.
} 
\label{fig:failure_case_analysis}
\vspace{-8pt}
\end{figure*}

\section{Why Small VLMs Fall Behind?}
\label{sec:problem_analysis}

We start with the InternVL series of models. On RefCOCO benchmarks, we randomly select a subset and visualize the predictions from the 8B VLM alongside the ground truth bounding boxes. 
Following previous work~\cite{zhu2025internvl3,wang2025internvl35,bai2025qwen25}, we define the model $M_\theta$ \textit{succeeds} on a sample $d_i$ if the Intersection-over-Union (IoU) between the bounding box predicted by the model $M_\theta$ and the ground truth is higher than 0.5. The \textit{grounding success} is defined as 
\begin{equation}
\label{eq:success}
    \mathrm{S}_{\theta,i} =
    \begin{cases}
        1, & \text{if } \mathrm{IoU}_{\theta,i} > 0.5, \\[2pt]
        0, & \text{otherwise.}
    \end{cases}
\end{equation}

We observe that a large proportion of the failure cases of the small VLM is due to \failreason: the text prompt is semantically complex and there are multiple similar candidates in the image, so the model mis-locates to another candidate rather than the target object.
For example, in the top left example of Figure~\ref{fig:failure_case_analysis}, the model correctly understands the semantics of ``chair'', but ignores the instructions ``empty'' and ``next to guy''.

\begin{table}[t]
\centering
\setlength{\tabcolsep}{10pt}
\caption{\textbf{Analysis of failure cases by proprietary models.} Reasons identified by different commercial models based on failure cases generated from small VLMs ($\text{IoU} < 0.5$). A large proportion of failure cases is due to \failreason.}
\vspace{-6pt}
\begin{tabular}{lccc}
\toprule
\textbf{Failure Reason} & \textbf{GPT-4} & \textbf{GPT-5} & \textbf{Gemini-3-Pro} \\
\midrule
COMPLEX-PROMPT & 62.8\% & 48.8\% & 46.8\% \\
\bottomrule
\end{tabular}

\label{tab:failure_error_distribution}
\vspace{-2pt}
\end{table}

\input{figures/complex_prompt_resolve.tex}

To further verify, we collect a set of failure cases and analyze how these cases improve with model size. 
Specifically, we collect the failure cases from InternVL-3-8B models then send these images together with ground truth to a proprietary model, and ask it to analyze the reason for all failure cases where the IoU between the prediction and ground truth boxes is lower than $0.5$. The full prompt to the proprietary model is provided in Appendix~\ref{app:prompt_failure_case_analysis}.
We categorize the failure cases into five categories:`COMPLEX-PROMPT', `AMBIGUOUS-IMAGE', `SMALL-OBJECT', `GT-ERROR', `OTHER', and judge with different models to provide a fair judgement.
Table~\ref{tab:failure_error_distribution} shows the key analysis using different proprietary models: Among all failure cases, all the proprietary models agree that a large proportion of failure cases (52.8\% on average) is due to \failreason, which confirms our observation. 
We attach the full failure reasons percentages in Appendix~\ref{app:detail_analysis_failure_case_small_vlm}.

We further compare with the predictions of bigger models and find that the \failreason~failure cases of the 8B model can be gradually resolved as the model size grows, as shown in Columns 2 and 3 of Table~\ref{tab:complex_prompt_resolve}. 
This motivates us to introduce reasoning to small models to mitigate the gap in understanding \failreason~compared to bigger models.

%% file: figures/complex_prompt_resolve.tex
\begin{table}[t]
\centering
\setlength{\tabcolsep}{6pt}
\caption{\textbf{Performance of different models on the \failreason subset.} We report the performance on the subset of RefCOCO benchmarks where InternVL-3-8B fails due to \failreason. `Acc' denotes the grounding success ratio.}
\vspace{-6pt}
    \begin{tabular}{l cc ccc c}
    \toprule
    \multirow{2}{*}{\textbf{Model}} & \multicolumn{2}{c}{InternVL-3} & \multicolumn{3}{c}{Gemini} & \textbf{Ours} \\
    \cmidrule(lr){2-3} \cmidrule(lr){4-6} \cmidrule(lr){7-7}
     & 32B & 78B & 2.5-Flash & 2.5-Pro & 3-Pro & EGM-InternVL-3-8B \\
    \midrule
    \textbf{Acc} & 29.9\% & 42.6\% & 20.9\% & 22.1\% & 57.3\% & 42.7\% \\
    \bottomrule
    \end{tabular}
    
\label{tab:complex_prompt_resolve}
\vspace{-2pt}
\end{table}

%% file: sec/4_method.tex
\section{Method}
\label{sec:method}

This section presents our method to equip small models with multi-modal reasoning capability through extended test-time computation. More specifically, we train the model with SFT to obtain the thinking pattern, followed by RL training~\cite{guo2025deepseek}. We describe the data curation for SFT in Section~\ref{sec:sft_data_curation}, the data curation for RL in Section~\ref{sec:rl_data_curation}, and the training pipeline in Section~\ref{sec:training_pipeline_and_reward}. Beyond the standard vanilla grounding task, we extend our method to the amodal grounding. Note that for \textbf{vanilla} grounding and \textbf{amodal} grounding, we curate different training datasets, and train the models \textbf{separately} for these two different tasks on the two different training datasets. Statistics of the curated training datasets are summarized in Table~\ref{tab:dataset_statistics}. We use proprietary VLMs in the data curation pipeline, and the corresponding prompts are provided in the Appendix~\ref{app:detailsftdata}.

\input{figures/dataset_statistics.tex}

\begin{figure}[t]
	\centering
\includegraphics[width=0.9\linewidth]{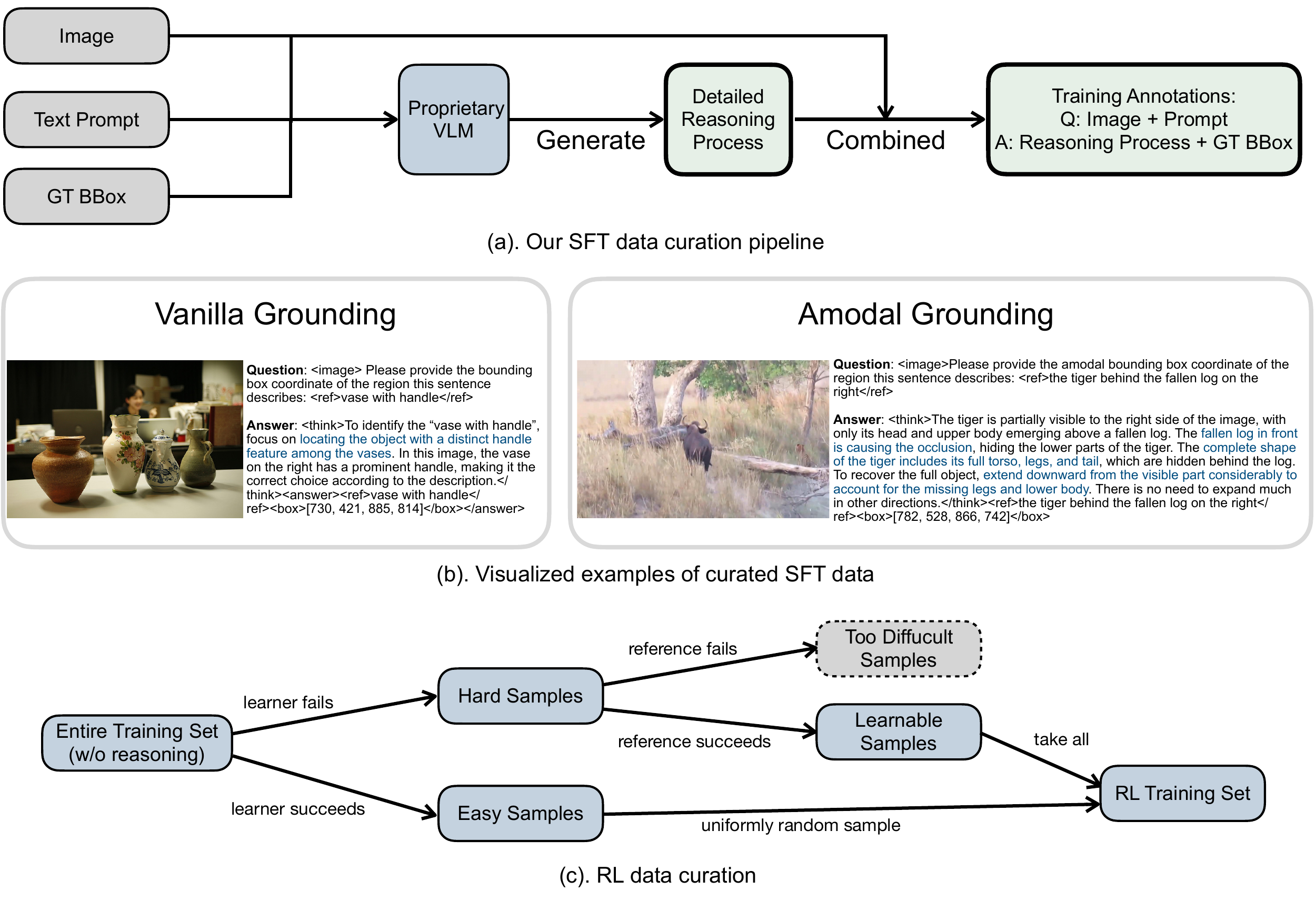}
\caption{
\textbf{Overview of our method.} 
\emph{Top (a): Data curation pipeline of SFT training data with reasoning.} We feed the image, text prompt and ground truth bounding box of the target object into a proprietary VLM to generate the detailed reasoning process of how to locate the object correctly given the image and text prompt. The generated reasoning process is incorporated as part of the training data.
\emph{Middle (b): Examples of generated reasoning training data for vanilla grounding and amodal grounding.} The reasoning process of vanilla grounding analyzes the feature that distinguishes the target object from others, and the reasoning process of amodal grounding further involves what object causes the occlusion and in which directions the visible parts should be extended to recover the complete object.
\emph{Bottom (c): RL data curation for vanilla grounding.} The RL data is curated by collecting the instances with learnability > 0 (\emph{i.e.,} the learner model fails while the reference model succeeds) and merging with \textit{easy samples} where the learner model succeeds.
} 
\label{fig:method}
\vspace{-8pt}
\end{figure}

\subsection{SFT Data Curation}
\label{sec:sft_data_curation}
The objective of the SFT training stage is to teach the model to learn the reasoning pattern in grounding tasks. As analyzed in Section~\ref{sec:problem_analysis}, addressing the key failure pattern requires reasoning about the how to locate the targe object accurately.

\vspace{2pt}
\noindent \textbf{Vanilla Grounding.}  
Given an off-the-shelf vanilla grounding training dataset $\mathcal{D}_{G} = \{d_i\}_{i=1}^N$, each sample $d_i$ contains an image, a textual query, and the corresponding ground-truth bounding box. We present each $d_i$ to a proprietary VLM~$\Phi$ to obtain its reasoning trace:
\begin{equation}
    z_i = \Phi(d_i).
\end{equation}
We then pair each original sample $d_i$ with its reasoning trace $z_i$ to obtain the reasoning training dataset $\mathcal{D}_C = \{ (d_i, z_i) \}_{i=1}^N$. The process of SFT data curation is illustrated on the top of Figure~\ref{fig:method}. Dataset statistics for $\mathcal{D}_{G}$ and $\mathcal{D}_{C}$ are reported in Table~\ref{tab:dataset_statistics} as the vanilla grounding category.

An example of vanilla grounding is shown on the middle left of Figure~\ref{fig:method}. For the text prompt `vase with handle', the answer involves a reasoning process that analyzes the distinct feature of the target vase compared with others.

\vspace{2pt}
\noindent \textbf{Amodal Grounding.} 
Unlike vanilla grounding, there is no off-the-shelf data for training that provides both the text prompt and the corresponding amodal bounding box. 
To bridge the gap, we use a proprietary VLM to generate a more detailed description of the target object that can uniquely specify it in the existing amodal segmentation datasets, such as COCO-Amodal~\cite{zhu2017semantic} and TAO-Amodal~\cite{hsieh2023tracking}. We further conduct a verification step on the generated prompts to remove the noisy ones. 
We denote this generated dataset as $\mathcal{D}_{T}$.

Once we obtain the text prompt associated with the amodal bounding box, we next generate the reasoning process via the same pipeline. The only difference is that, in addition to locating the target object, the reasoning should also involve: 
1) what object(s) are causing the occlusion; 
2) what this object's complete unoccluded shape looks like; 
3) in which direction(s) and how much the visible parts should be extended to recover the complete object. 
An example of generated reasoning data for amodal grounding is shown on the middle right of Figure~\ref{fig:method}. The task is to predict the amodal bounding box of the tiger behind the fallen log on the right. The reasoning process thus involves `the fallen log in front is causing the occlusion', `the complete shape of the tiger includes its full torso, legs, and tail', and `to recover the full object, extend downward from the visible part considerably'. This process yields the final amodal grounding dataset with reasoning $\mathcal{D}_{A}$. Additional details are provided in Appendix~\ref{app:detailsftdata}.

\subsection{RL Data Curation}
\label{sec:rl_data_curation}

\noindent \textbf{Vanilla Grounding.} For RL training, we want to select harder but learnable samples to enable the model to learn the grounding capability. More specifically, we select a harder subset of the off-the-shelf vanilla grounding training dataset $\mathcal{D}_{G}= \{d_i\}_{i=1}^{N}$ in Table~\ref{tab:dataset_statistics} for vanilla grounding RL training by learnability~\cite{evans2024data,zhan2025elip}. 

More specifically, given the learner model $M_{\theta}$ and the reference model $M_{\text{ref}}$, 
we first compute their IoU values on the entire training dataset $\mathcal{D}_{G} $, then we know whether a model \textit{succeeds} on each sample as in Equation~\ref{eq:success}. 
The learnability of each sample $d_i$ is calculated as 
$l_i = \mathrm{S}_{\text{ref},i} - \mathrm{S}_{\theta,i}$. The learner model is a small VLM that we train, and the reference model is a state-of-the-art visual grounding model.
For each learner model, we then construct a learnable sample set 
$\mathcal{D}_{\text{learnable}} = \{\, d_i \mid l_i > 0,\, \, d_i \in \mathcal{D}_{G} \,\}$, where the learnability is greater than 0. This means that the learner model fails on these samples while the reference model succeeds on these samples. Therefore, these samples have not yet been learned by the learner model, while they are learnable as the reference model can succeed on them. The rest samples where the reference model also fails are too difficult or may be due to ground truth annotation errors and we thus do not use them.

If we only use samples where the learner model fails for RL training, the model may overfit on harder samples and the performance for general cases can degrade.
Therefore, we also uniformly sample instances where the learner model succeeds from $\mathcal{D}_{G}$ to form an easy sample set $\mathcal{D}_{\text{easy}}$ of equal size to $\mathcal{D}_{\text{learnable}}$. 
The final training dataset is the union of the two subsets:
$\mathcal{D} = \mathcal{D}_{\text{learnable}} \cup \mathcal{D}_{\text{easy}}.$

\vspace{2pt}
\noindent \textbf{Amodal Grounding.} We use the entire amodal grounding dataset $\mathcal{D}_T$ in Table~\ref{tab:dataset_statistics} to make the most of the limited data and to strengthen the model's performance in this new task.

\subsection{Training Pipeline and Reward}
\label{sec:training_pipeline_and_reward}

\vspace{2pt}
\noindent \textbf{Training Pipeline.} 
We first perform SFT on the base model to teach it the desired reasoning pattern. Then, we train the model using reinforcement learning to obtain the final \method~models. We adopt GRPO~\cite{shao2024deepseekmathpushinglimitsmathematical} with a token-level mean loss~\cite{yu2025dapoopensourcellmreinforcement}. We introduce KL divergence penalty and entropy regularization to ensure the model explores sufficiently during reasoning while avoiding pattern collapse.

Specifically, for a given input $q$ and its $i$-th generated sequence $o_i$ sampled from the behavior policy 
$\pi_{\theta_{\mathrm{old}}}$, we denote its length by $|o_i|$ and its token sequence as $\{o_{i,t}\}_{t=1}^{|o_i|}$. 
The overall GRPO objective is defined as:
{\small
\begin{align}
\mathcal{J}_{\mathrm{GRPO}}(\pi_\theta)
&= 
\mathbb{E}_{q\sim P(Q),\,\{o_i\}_{i=1}^{G}\sim\pi_{\theta_{\mathrm{old}}}(O\mid q)} \\
&\Bigg[
\frac{1}{G}\sum_{i=1}^{G}
\sum_{t=1}^{|o_i|}
\min\!\Bigg(
\frac{\pi_{\theta}(o_{i,t}\mid q,o_{i,<t})}
     {\pi_{\theta_{\mathrm{old}}}(o_{i,t}\mid q,o_{i,<t})}
\,\hat{A}_{i,t},
\notag\\[-1mm]
&
\mathrm{clip}\!\left(
\frac{\pi_{\theta}(o_{i,t}\mid q,o_{i,<t})}
     {\pi_{\theta_{\mathrm{old}}}(o_{i,t}\mid q,o_{i,<t})},
1-\epsilon,\,1+\epsilon
\right)\!
\hat{A}_{i,t}
\Bigg)
\notag\\[-1mm]
&\qquad
-\;\beta\,\mathbb{D}_{\mathrm{KL}}\!\big(\pi_\theta\,\|\,\pi_{\mathrm{ref}}\big)
+\;\gamma\,\mathbb{H}\!\big(\pi_\theta\big)
\Bigg],
\end{align}
}
\begin{align}
\mathbb{D}_{\mathrm{KL}}\!\big(\pi_\theta\|\pi_{\mathrm{ref}}\big)
&=
\frac{\pi_{\mathrm{ref}}(o_i\mid q)}{\pi_\theta(o_i\mid q)}
- \log\frac{\pi_{\mathrm{ref}}(o_i\mid q)}{\pi_\theta(o_i\mid q)} - 1,\\
\mathbb{H}\!\big(\pi_\theta\big)
&= - \log \pi_\theta(o_i\mid q),
\end{align}
where $\beta$ and $\gamma$ are hyper-parameters; $\epsilon$ is PPO clip ratio; $\pi_{\mathrm{ref}}$ is the reference model; and $\hat{A}_{i,t}$ is the token level advantage, derived
 from the rewards $\{r_1,r_2,.. .,r_G\}$ corresponding to the outputs within each group:
\begin{equation}
\hat{A}_i
= \frac{r_i - \mathrm{mean}\!\big(\{r_j\}_{j=1}^{G}\big)}
        {\mathrm{std}\!\big(\{r_j\}_{j=1}^{G}\big)}, \hat{A}_{i,t}=\hat{A}_i/|o_i|.
\end{equation}

\vspace{2pt}
\noindent \textbf{Reward.} 
We define the reward as a weighted combination of the IOU between the predicted bounding box and ground truth, and \textit{grounding success} (as in Equation~\ref{eq:success}) with hyper-parameter $\alpha$ :  
\begin{equation}
\label{eq:reward}
    r_i = \alpha \, IoU_{\theta,i} + (1-\alpha)\,\mathrm{S}_{\theta,i}.
\end{equation}
The $\alpha$ is set to 0.5 by default during experiments unless specifically clarified.

%% file: figures/dataset_statistics.tex
\begin{table*}[t]
\centering
\setlength{\tabcolsep}{5pt}
\renewcommand{\arraystretch}{1.1}
\caption{
\textbf{Statistics of curated training datasets} for vanilla grounding and amodal grounding. 
The vanilla grounding is curated from RefCOCO training dataset~\cite{liu2025nvila} and amodal grounding dataset is curated from COCO-Amodal~\cite{zhu2017semantic} and TAO-Amodal~\cite{hsieh2023tracking}.
}
\vspace{-6pt}
\resizebox{\linewidth}{!}{
\begin{tabular}{c c c c c c}
\toprule
Category & \#Samples & \#Images & Avg. \#BBoxes  & Width & Height  \\
\midrule
Vanilla Grounding & 575,208 & 24,407 & $\text{2.3}_{\pm 1.3}$ & $\text{585.7}_{\pm 87.0}$ & $\text{480.3}_{\pm 93.1}$ \\
Amodal Grounding  & 23,698 & 15,798 & $\text{1.5}_{\pm 1.5}$ & $\text{1186.2}_{\pm 332.0}$ & $\text{717.2}_{\pm 175.2}$\\

\bottomrule
\end{tabular}
}
\label{tab:dataset_statistics}
\vspace{-2pt}
\end{table*}

%% file: sec/5_experiment.tex
\section{Experiments}
\label{sec:experiments}

\subsection{Implementation Details}
\label{sec:exp_implementation_detail}

\vspace{2pt}
\noindent \textbf{Models.} We apply our method to the state-of-the-art visual grounding open-source VLM families InternVL and QwenVL, and add a prefix `\method-' to models to which our method has been applied. Our experiments are conducted on InternVL-3-1B, InternVL-3-2B and InternVL-3-8B for the InternVL family, and on Qwen3-VL-2B-Thinking, Qwen3-VL-4B-Thinking and Qwen3-VL-8B-Thinking for the QwenVL family.

\vspace{2pt}
\noindent \textbf{SFT.} For the SFT training stage, we use the official GitHub scripts of InternVL-3 and Qwen3-VL. The training dataset statistics are presented in Table~\ref{tab:dataset_statistics}. The learning rate is set to be $1e^{-5}$, the epoch is set to be $1$ and the training batch size is $128$. We train all models on 8 A100 GPUs.

\vspace{2pt}
\noindent \textbf{RL.} For the reinforcement learning stage, we employ the VeRL framework~\cite{sheng2024hybridflow}. All rollouts and inference are performed using the vLLM engine~\cite{kwon2023efficient}. We train with a learning rate of $3e^{-6}$ for $5$ epochs and a batch size of $256$. The reward weight $\alpha$ in Equation~\ref{eq:reward} is fixed to $0.5$. For the QwenVL model series, we set the KL coefficient to $\beta=0.005$ and the entropy coefficient to $\gamma=0.0$. For the InternVL model series , we use $\beta=0.0$ and $\gamma=0.01$. We construct the RL training dataset following the procedure described in Section~\ref{sec:rl_data_curation}, using InternVL-3.5-241B as the reference model $M_{\text{ref}}$ and the corresponding fine-tuned model as the learner model $M_{\theta}$. 
The final dataset size varies across models. The full details about our RL training and dataset samples are attached in Appendix~\ref{app:detailrl}.

\subsection{Evaluation Benchmarks and Metrics}
\label{sec:evalbenchmark}

\noindent \textbf{Vanilla Grounding Benchmark.} For vanilla grounding, we use the standard RefCOCO~\cite{kazemzadeh2014referitgame,mao2016generation} benchmarks for evaluation, which contain eight test splits. We adopt `accuracy' as the evaluation metric, \emph{i.e.,} the proportion of \textit{grounding success} $S_{\theta,i}$ as defined in Equation~\ref{eq:success}. We report the score on individual splits as well as the average score over the eight splits.

\vspace{2pt}
\noindent \textbf{Amodal Grounding Benchmark.} As there is no off-the-shelf benchmark for amodal grounding, we start from the standard amodal benchmark in the computer vision community -- the val and test splits of COCO-Amodal~\cite{zhu2017semantic} dataset, and use the pipeline as introduced in Section~\ref{sec:sft_data_curation} to generate text prompts. We therefore obtain an evaluation benchmark consisting of 11,261 samples and 2,474 images. We adopt the same evaluation metric as vanilla grounding.

\vspace{2pt}
\noindent \textbf{Efficiency Metric.} To measure efficiency, besides the number of model parameters, we also calculate the average latency per sample on a random 10\% subset of the standard RefCOCO benchmarks for evaluation.
Latency is measured as the time (in milliseconds) from sending the text prompt to receiving the entire answer from the model. For fair comparison, we report the total \emph{GPU latency} on L20 GPUs.

\vspace{2pt}

\noindent  We adopt identical inference hyper-parameters for all models during evaluation. The only differences in the prompts follow the model-family-specific formats prescribed in the official implementation. 
Further details of the evaluation setup are provided in Appendix~\ref{app:detaileval}.

\subsection{Vanilla Grounding}
\label{sec:exp_vanilla_grounding}

\input{figures/vanilla_grounding.tex}

Table~\ref{tab:vanilla_grounding_accuracy} shows the results of applying our method to models of different sizes from the InternVL-3 family and the Qwen3-VL family. We show the ablation of SFT and RL in Appendix~\ref{app:ablation_sft_rf}.

\begin{sloppypar}
\vspace{2pt}
\noindent\textbf{\method~consistently improves the grounding performance.} 
\method boosts the performance of Qwen3-VL-2B/4B/8B-Thinking by +6.0/+3.8/+3.6 and InternVL-3-1B/2B/8B by +5.2/+1.7/+1.1 for the average accuracy. This shows the generality of our \method that can improve  small VLMs of \textit{different sizes} and from \textit{different families}
on RefCOCO benchmarks.
\end{sloppypar}

\begin{sloppypar}
\vspace{2pt}
\noindent\textbf{\method is mainly boosted by reasoning.} 
(1) The ablation study that disables reasoning (\(+\method\) w/o \(R\)) in Table~\ref{tab:vanilla_grounding_accuracy}, with full results reported in Appendix~\ref{app:ablation_reasoning_process}, verifies the necessity of the reasoning process, as EGM performance drops notably without it. 
(2) The commercial models actually perform worse (last section of Table~\ref{tab:vanilla_grounding_accuracy}) compared with all \method models; it is difficult to distill from an inferior model to boost performance.

\end{sloppypar}

\begin{sloppypar}
\vspace{2pt}
\noindent\textbf{\method mitigates the gap between small models and larger models.} For example, the performance of Qwen3-VL-2B/4B/8B-Thinking is 83.6/87.2/87.8. Our 2B and 4B models outperform Qwen3-VL-8B-Thinking by +1.8/+3.2. 
Our \method shows that extending the test-time computation can boost the performance by improving text understanding. Furthermore, the fine-grained analysis in Appendix~\ref{app:hard_sample} confirms that our method is particularly effective on hard samples, resolving nearly half of the cases with baseline IoU$<0.5$. 
\end{sloppypar}

\begin{sloppypar}
\vspace{2pt}
\noindent\textbf{\method is faster and more accurate than Qwen3-VL-Thinking.} For the Qwen3-VL models, 2B/4B/8B-Thinking does not outperform 2B/4B/8B-Instruct (-0.6/-0.1/-0.5), but our 2B/4B/8B reasoning models significantly outperform 2B/4B/8B-Instruct (+5.4/+3.7/+3.1), while being faster than 2B/4B/8B-Thinking, demonstrating the effectiveness and efficiency of our multi-modal reasoning.
\end{sloppypar}

\subsection{Amodal Grounding}
\label{sec:exp_amodal_grounding}

\input{figures/amodal_grounding.tex}

To further verify the effectiveness of our method, we show the accuracy results on the new task of amodal grounding in Table~\ref{tab:amodal_grounding_accuracy}. Similar to vanilla grounding, \textbf{our method consistently boosts the performance of models of different sizes and from different families} by +7.4/+6.4/+11.5 (InternVL-3) and +6.9/+2.3/+2.5 (Qwen3-VL), demonstrating the wide applicability of our method for different tasks.

\begin{figure*}[t]
	\centering
\includegraphics[width=1.0\linewidth]{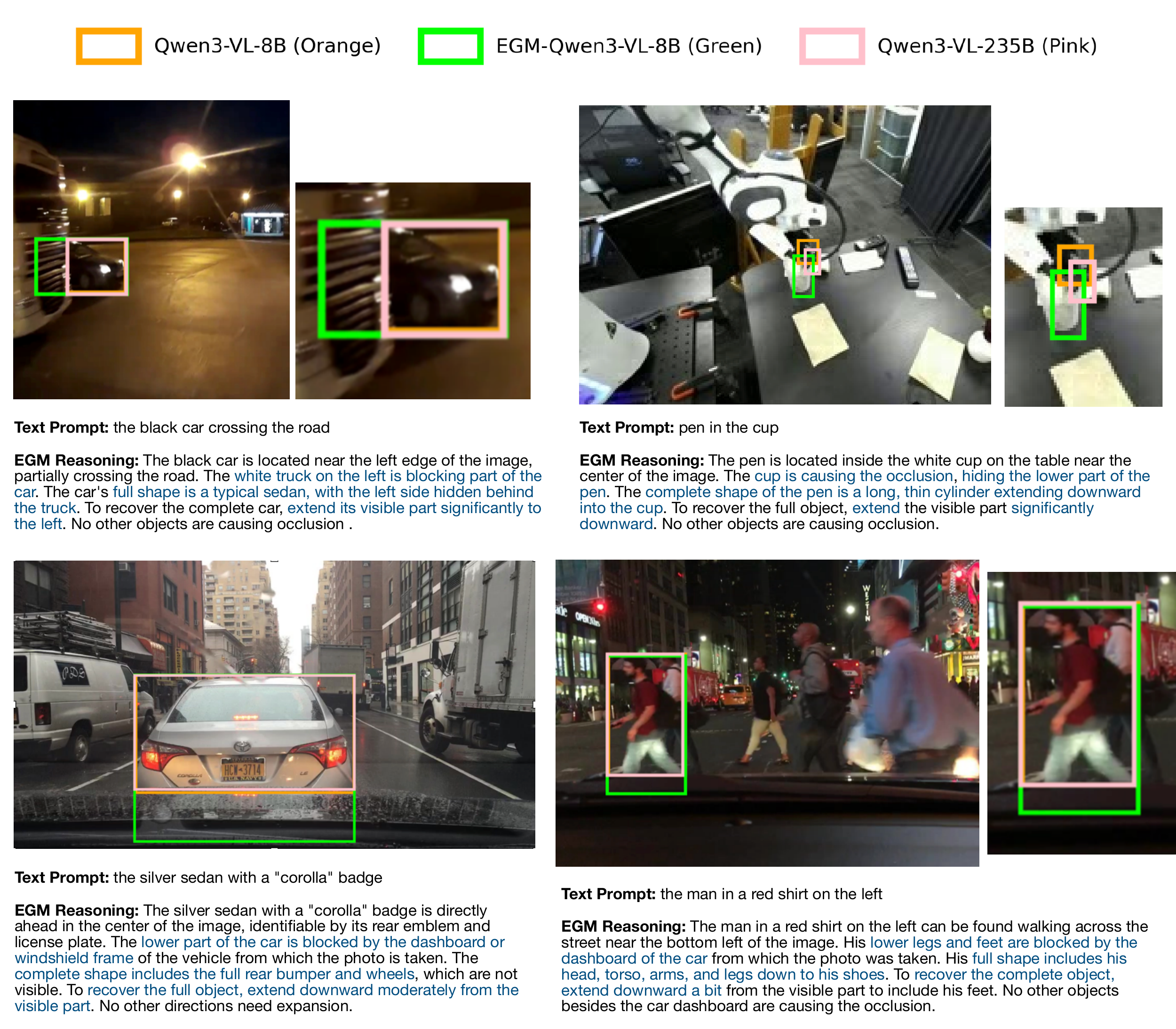}
\caption{
\textbf{Demo} of Qwen3-VL-8B, Qwen3-VL-235B, and our \method-Qwen3-VL-8B for amodal grounding in autonomous driving and robotics scenarios.
} 
\label{fig:demo_driving_robotics}
\end{figure*}

Moreover, in Figure~\ref{fig:demo_driving_robotics} we show demos of our \method amodal grounding models in autonomous driving and robotics scenarios. 
The top left demo is a situation called `ghost probe', where a car crossing the road suddenly appears. In this situation, it is important to notice and figure out the complete shape of the occluded car when they have not fully appeared. 
The top right demo for robotics is collected from the DROID dataset~\cite{khazatsky2024droid}. In this situation, the robot arm is going to manipulate the objects on the table and it is important to know the complete shape of occluded objects, \emph{e.g.,} the pen, for robust manipulation, as discussed in~\cite{xia2024targo}.
The bottom demos are two additional demos for driving scenario.
For all cases, it can be observed that compared with Qwen3-VL-8B and Qwen3-VL-235B, our \method-Qwen3-VL-8B can reason about the occlusion situations and correctly predict the complete shape of the occluded pedestrian, the occluded car and the occluded pen.
This suggests the potential application of our \method models in autonomous driving and robotic manipulation tasks.

\begin{figure}[ht]
\captionsetup{aboveskip=-2pt}
    \centering
    \includegraphics[width=0.97\linewidth]{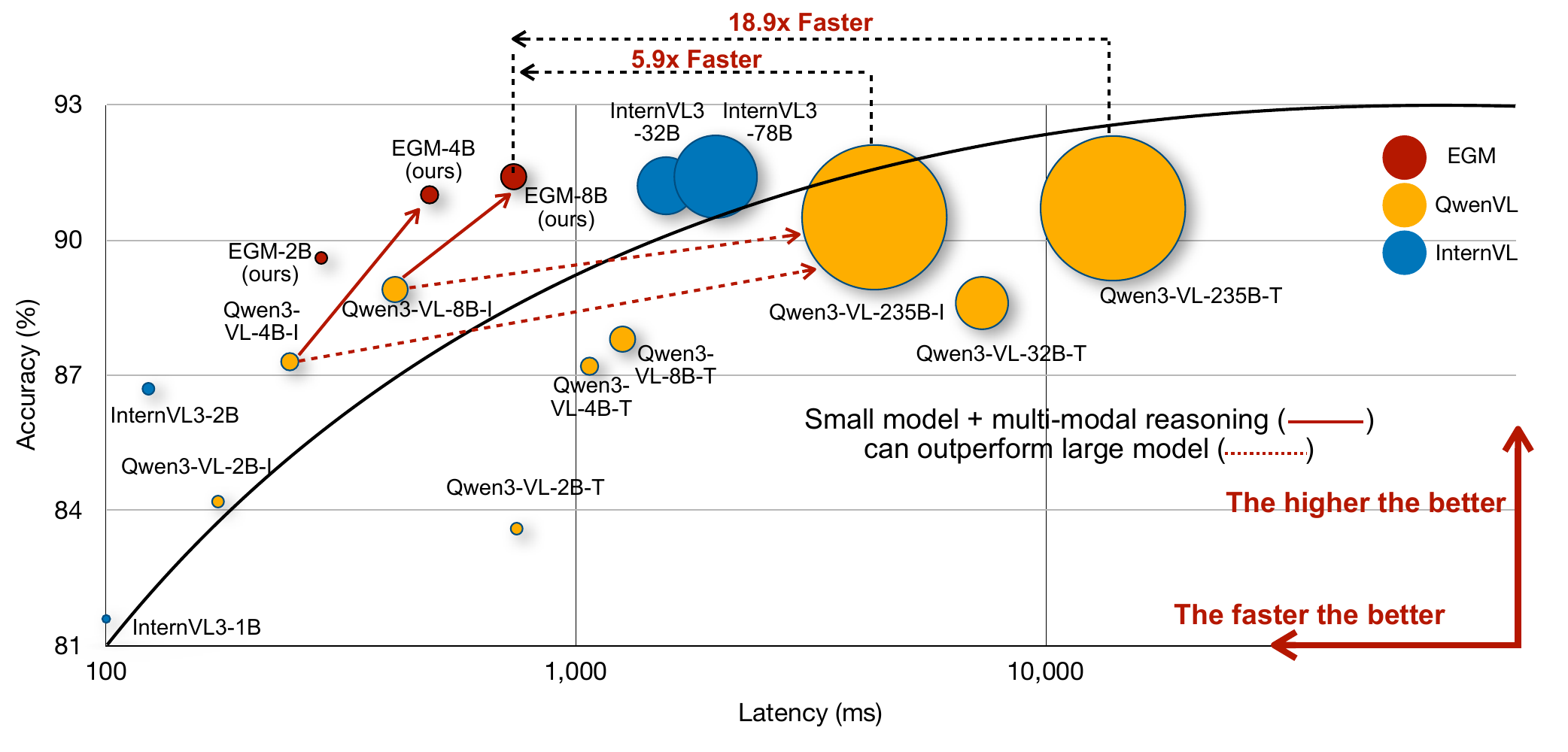}
    \caption{\textbf{Accuracy vs. Efficiency.} Our models, such as \method-Qwen3-VL-4B and \method-Qwen3-VL-8B, have greatly improved the efficiency of visual grounding. For example, \method-Qwen3-VL-8B outperforms both the state-of-the-art Qwen3-VL-235B-Instruct and Qwen3-VL-235B-Thinking models for accuracy, while speeding up 5.9$\times$/18.9$\times$ in terms of GPU latency. For Qwen models, `-T' denotes `-Thinking' and `-I' denotes `-Instruct'.
    }
    \label{fig:efficiency_tradeoff}
    \vspace{-8pt}
\end{figure}

\begin{figure}[ht]
	\centering
\includegraphics[width=0.9\linewidth]{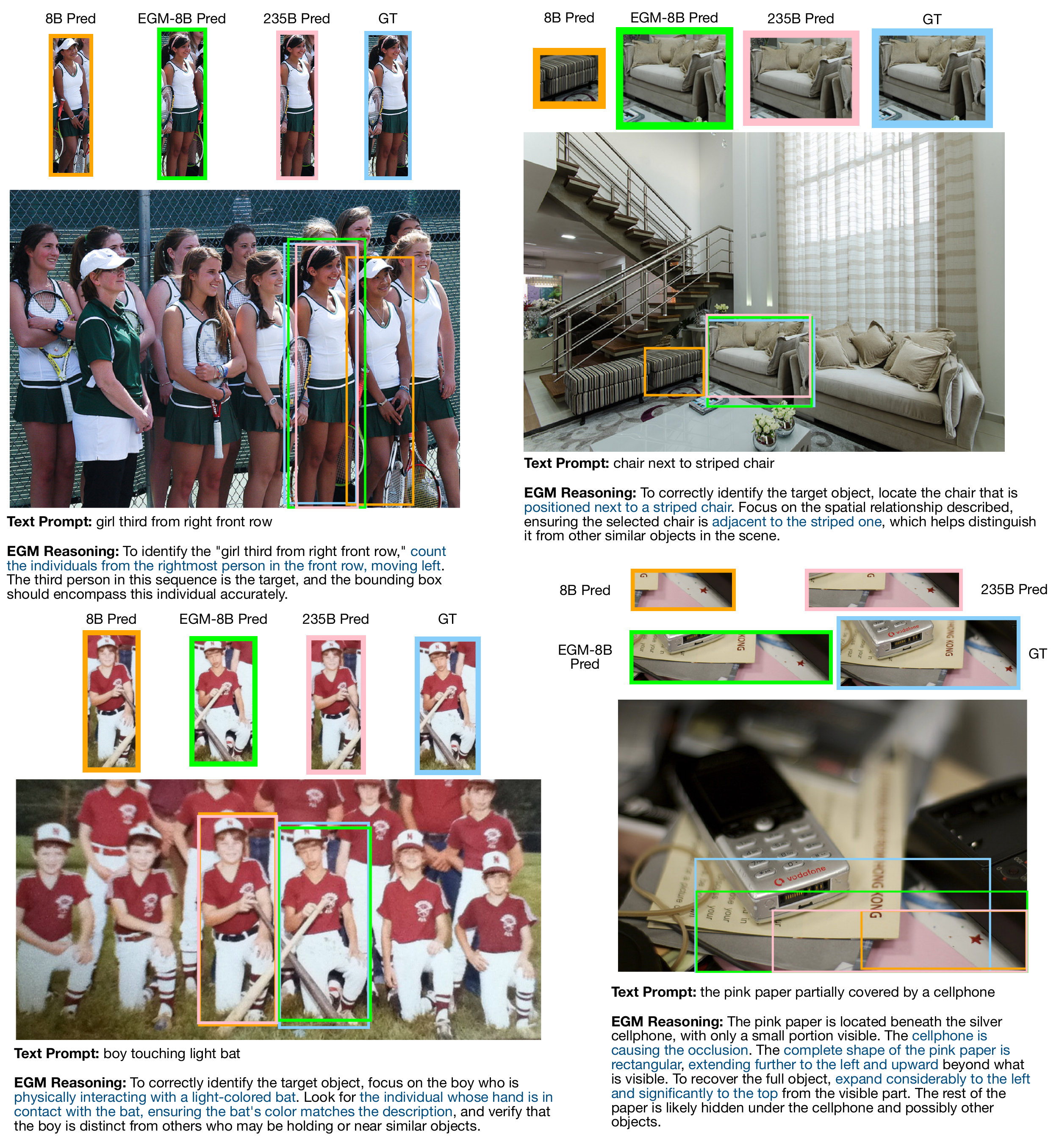}
\caption{
\textbf{Qualitative comparison} of Qwen3-VL-8B, Qwen3-VL-235B, and our \method-Qwen3-VL-8B for vanilla grounding and amodal grounding. Vanilla grounding: top left, top right and bottom left; Amodal grounding: bottom right.
} 
\label{fig:qualitative_qwen3}
\vspace{-8pt}
\end{figure}

\subsection{Efficiency Comparison}
\label{sec:exp_efficiency_comparison}

Figure~\ref{fig:efficiency_tradeoff} compares the accuracy and efficiency of models in the QwenVL family. \textbf{Our \method models significantly improve the efficiency of visual grounding}: Both our 4B and 8B models surpass the 235B models in accuracy and are over 5$\times$ faster in inference. Our 2B model also outperforms several larger models—32B-Thinking, 8B-Instruct, and 8B-Thinking—while achieving better GPU latency. The observation is consistent with the results in Table~\ref{tab:vanilla_grounding_accuracy}, demonstrating the effectiveness of extending test-time computation for visual grounding.

\subsection{Qualitative Comparison}
\label{sec:exp_qualitative}

Figure~\ref{fig:qualitative_qwen3} shows the qualitative comparison of Qwen3-VL-8B, Qwen3-VL-235B, and our \method-Qwen3-VL-8B for vanilla grounding and amodal grounding. Please refer to Appendix~\ref{app:qualitative_results} for more qualitative results.

The top left, top right and bottom left samples are vanilla grounding on RefCOCO benchmark. For the \emph{top left} example, the text prompt is `girl third from right front row', and the Qwen3-VL-8B model mis-locates to the `second' girl. Our \method-Qwen3-VL-8B correctly reasons the key feature `count the individuals from the rightmost person in the front row, moving left' and generates a correct prediction. For the \emph{top right} example, the text prompt is `chair next to striped chair', and the Qwen3-VL-8B model confuses it with the striped chair itself. Our \method-Qwen3-VL-8B, however, correctly reasons about the key feature `the chair that is positioned next to a striped chair' and outputs the correct grounding box. For the \emph{bottom left} example, the text prompt is `boy touching light bat', and the Qwen3-VL-8B model fails to understand `touching light bat' and mis-predicts the boy touching the darker bat. Our \method-Qwen3-VL-8B predicts the correct boy, with a reasoning that it should be `the individual whose hand is in contact with the bat, ensuring the bat's color matches the description'. For the vanilla grounding cases, generally Qwen3-VL-235B performs better than Qwen3-VL-8B, as can be observed in the examples.

The \emph{bottom right} example is amodal grounding on the COCO-Amodal benchmark. It can be observed that Qwen3-VL-8B and Qwen3-VL-235B fail to predict the amodal grounding box of `the pink paper partially covered by a cellphone', while our \method-Qwen3-VL-8B prediction is close to the ground truth, with reasoning process about locating the pink paper, analyzing it is occluded by the cellphone and other objects and should extend to the left and upward to recover the full unoccluded shape.
This reasoning helps \method-Qwen3-VL-8B to better locate the amodal bounding box covering both the visible and occluded parts of the pink paper.

%% file: figures/vanilla_grounding.tex
\begin{table*}[t]
    \centering
    \setlength{\tabcolsep}{5pt}
    \renewcommand{\arraystretch}{1.2}

    \caption{\textbf{Accuracy results for vanilla grounding.} Our method ($+\method$) consistently improves the performance of models of different sizes and from different families. The ablation study of reasoning process ($+\method\ w/o\ R$) validates the effectiveness of our method.
    Entries marked with * are from the official report~\cite{wang2025internvl35}, and the others are measured by ourselves.
    }
    \vspace{-6pt}

    \resizebox{\linewidth}{!}{
    \begin{tabular}{l c c c c c c c c l}
    \toprule[1.5pt]
    \multirow{2}{*}{Model} & \multicolumn{3}{c}{RefCOCO} & \multicolumn{3}{c}{RefCOCO+} & \multicolumn{2}{c}{RefCOCOg} & \multirow{2}{*}{Avg. Acc} \\
     & val & test-A & test-B & val & test-A & test-B & val & test &  \\
     \midrule[1.2pt]
    Qwen3-VL-2B-Instruct & 88.7 & 90.8 & 84.7 & 80.1 & 85.7 & 73.1 & 84.8 & 85.7 & 84.2 \\
      Qwen3-VL-2B-Thinking & 87.5 & 91.4 & 82.0 & 80.2 & 86.7 & 72.1 & 84.4 & 84.5 & 83.6 \\
      \rowcolor{black!10} \quad + \textit{\method}  & 93.0 & 94.0 & 89.4 & 87.8 & 91.7 & 82.7 & 88.6 & 89.3 & ${\text{89.6}}_{\color{yellowgreen}+6.0\uparrow}$ \\
    \cmidrule(lr){1-10}
      Qwen3-VL-4B-Instruct & 91.0 & 92.7 & 87.6 & 84.4 & 88.7 & 78.2 & 88.3 & 87.8 & 87.3 \\
      Qwen3-VL-4B-Thinking & 90.0 & 92.7 & 85.6 & 85.2 & 89.5 & 79.3 & 87.8 & 87.7 & 87.2 \\
      \rowcolor{black!10} \quad + \textit{\method}  & 93.5 & 95.1 & 90.9 & 89.7 & 93.1 & 84.9 & 90.4 & 90.8 & ${\text{91.0}}_{\color{yellowgreen}+3.8\uparrow}$ \\
    \cmidrule(lr){1-10}
      Qwen3-VL-8B-Instruct & 91.6 & 93.3 & 87.8 & 85.8 & 90.3 & 79.9 & 88.7 & 88.7 & 88.3 \\
      Qwen3-VL-8B-Thinking & 91.0 & 92.9 & 86.9 & 86.2 & 89.3 & 80.2 & 87.6 & 88.6 & 87.8 \\
      \rowcolor{black!10} \quad + \textit{\method w/o R}  & 92.2 & 93.6 & 89.2 & 85.7 & 90.2 & 80.0 & 88.6 & 89.3 & ${\text{88.6}}_{\color{yellowgreen}+0.8\uparrow}$ \\
      \rowcolor{black!10} \quad + \textit{\method}  & 93.9 & 95.0 & 91.2 & 90.1 & 93.3 & 85.9 & 90.4 & 91.2 & ${\textbf{\text{91.4}}}_{\color{yellowgreen}+3.6\uparrow}$ \\
      \midrule
    Qwen3-VL-235B-A22B-Instruct & 92.9 & 94.6 & 90.2 & 88.6& 92.4 & 84.2 & 90.3 & 90.8 & 90.5 \\
    Qwen3-VL-235B-A22B-Thinking & 92.9 & 94.1 & 90.6& 89.5& 92.5& 85.5 &90.4 &90.5 &90.7 \\
    \midrule[1.2pt]
    InternVL-3-1B & 85.8 & 90.1  & 81.7  & 76.6  & 84.1  & 69.2 & 82.8  & 82.6  & 81.6* \\
      \rowcolor{black!10} \quad + \textit{\method}  & 90.2  & 93.2 & 87.0 & 83.8 & 88.8 & 77.5 & 86.4 & 87.5 & ${\text{86.8}}_{\color{yellowgreen}+5.2\uparrow}$ \\
    \cmidrule(lr){1-10}
      InternVL-3-2B & 89.8 & 92.6  & 86.4  & 84.0  & 89.2 & 76.5  & 87.6  & 87.2 & 86.7* \\
      \rowcolor{black!10} \quad + \textit{\method}  & 92.2 & 94.0 & 87.4 & 85.6 & 91.2 & 79.2 & 88.5 & 88.7 & ${\text{88.4}}_{\color{yellowgreen}+1.7\uparrow}$ \\
    \cmidrule(lr){1-10}
      InternVL-3-8B & 92.5 & 94.6  & 88.0  & 88.2  & 92.5 & 81.8  & 89.6 & 90.0 & 89.6* \\
       \rowcolor{black!10} \quad + \textit{\method w/o R}  & 91.3 & 93.9 & 87.0 & 85.7 & 90.6 & 78.1 & 87.3 & 88.2 & ${\text{87.7}}_{\color{red}-1.9 \downarrow}$ \\
      \rowcolor{black!10} \quad + \textit{\method}  & 93.6 & 95.2 & 90.1 & 89.3 & 93.6 & 83.1 & 89.7 & 90.7 & ${\text{90.7}}_{\color{yellowgreen}+1.1\uparrow}$ \\
   \midrule
    InternVL-3-78B  & 93.4 & 95.4 & 90.3 & 90.1 & 93.8 & 85.3 & 91.5 & 91.5 & \textbf{91.4}* \\ 
     \midrule[1.2pt]
     Gemini-2.5-Pro & 66.8 & 66.8 & 69.2 & 58.2 & 58.2 & 60.6 & 65.0 & 65.7 &  63.8 \\
     Gemini-3-Pro & 87.3 & 86.4 & 86.4 & 80.3 & 79.7 & 81.8 & 89.9 & 89.6 & 85.2 \\
     GPT-5 & 42.6 & 46.0 & 39.3 & 37.4 & 40.1 & 35.2 & 39.1 & 41.0 & 40.1\\
    \bottomrule[1.5pt]
    \end{tabular}
    }

    \label{tab:vanilla_grounding_accuracy}
    \vspace{-4pt}
    \end{table*}

%% file: figures/amodal_grounding.tex
\begin{table}[h]
    \centering
    \caption{\textbf{Accuracy results for amodal grounding.} Similar to vanilla grounding, our method consistently boosts the performance of amodal grounding.}
    \vspace{-6pt}
    \resizebox{\linewidth}{!}{
        \setlength{\tabcolsep}{3pt}
        \renewcommand{\arraystretch}{1.2}
        \begin{tabular}{l cccc cccc ccc}
            \toprule
            \multirow{2}{*}{\textbf{Setting}} & \multicolumn{4}{c}{\textbf{InternVL-3}} & \multicolumn{4}{c}{\textbf{Qwen3-VL}} & \multicolumn{3}{c}{\textbf{Others}} \\
            \cmidrule(lr){2-5} \cmidrule(lr){6-9} \cmidrule(lr){10-12}
             & 1B & 2B & 8B & 78B & 2B & 4B & 8B & 235B & Gemini-2.5-Pro & Gemini-3-Pro & GPT-5 \\
            \midrule
            
            Original & 
            $56.9\vphantom{_{\color{yellowgreen}+0.0\uparrow}}$ & 
            $63.8\vphantom{_{\color{yellowgreen}+0.0\uparrow}}$ & 
            $62.0\vphantom{_{\color{yellowgreen}+0.0\uparrow}}$ & 
            $51.0\vphantom{_{\color{yellowgreen}+0.0\uparrow}}$ & 
            $66.5\vphantom{_{\color{yellowgreen}+0.0\uparrow}}$ & 
            $71.6\vphantom{_{\color{yellowgreen}+0.0\uparrow}}$ & 
            $71.4\vphantom{_{\color{yellowgreen}+0.0\uparrow}}$ & 
            $74.1\vphantom{_{\color{yellowgreen}+0.0\uparrow}}$ & 
            $52.3\vphantom{_{\color{yellowgreen}+0.0\uparrow}}$ & 
            $71.9\vphantom{_{\color{yellowgreen}+0.0\uparrow}}$ & 
            $24.7\vphantom{_{\color{yellowgreen}+0.0\uparrow}}$ \\
            
            + \textit{\method} & 
            ${\text{64.3}}_{\color{yellowgreen}+7.4\uparrow}$ & 
            ${\text{70.2}}_{\color{yellowgreen}+6.4\uparrow}$ & 
            ${\text{73.5}}_{\color{yellowgreen}+11.5\uparrow}$ & 
            -- & 
            ${\text{73.4}}_{\color{yellowgreen}+6.9\uparrow}$ & 
            ${\text{73.9}}_{\color{yellowgreen}+2.3\uparrow}$ & 
            ${\text{73.9}}_{\color{yellowgreen}+2.5\uparrow}$ & 
            -- & -- & -- & -- \\
            
            \bottomrule
        \end{tabular}
    }
    
    \label{tab:amodal_grounding_accuracy}
    \vspace{-2pt}
\end{table}

%% file: sec/6_conclusion.tex
\section{Conclusion}
\label{sec:conclusion}

\begin{sloppypar}
In this paper, we introduced \emph{`\textbf{E}fficient visual \textbf{G}rounding language \textbf{M}odels' (\method)}, a method to improve the efficiency of visual grounding language models. Our method equips small models with the multi-modal reasoning capability via a two-stage SFT-RL training paradigm. Experiments show our method can be applied to different model families, different model sizes, and different tasks, to boost the performance consistently and significantly.
Therefore, our \method~achieves a 91.4 IoU with an 8B model, outperforming the conventional 235B model with a 90.5 IoU, while speeding up 5.9$\times$.
We hope systems developed in this way can help scenarios such as autonomous driving and robotics, where an efficient grounding language model with amodal capability is important.
\end{sloppypar}

%% file: sec/7_supple.tex
\appendix
\renewcommand{\theHsection}{\Alph{section}}

\onecolumn
\begin{center}
{\LARGE\bfseries Appendix}
\end{center}
\vspace{-0.5em}

\section{Implementation Details}
In this section, we provide the detailed experimental settings referenced in Section~\ref{sec:exp_implementation_detail} and Section~\ref{sec:evalbenchmark} of the main paper. 
Specifically, we report the hyperparameters used for RL rollout and training in Table~\ref{tab:RLpara}, the sizes of the RL training datasets for different models in Table~\ref{tab:RLdata_num}, 
and the inference configurations in Table~\ref{tab:modelevalution}. 
We also present the prompt templates used for various models and tasks in Figure~\ref{fig:promptfortask}.
\subsection{Details of RL Training }
\label{app:detailrl}

\begin{table}[ht]
    \centering
    
    \caption{\textbf{Hyperparameters for RL training and rollout.} 
In the VeRL configuration, setting \texttt{top\_k}~\(=-1\) disables top-\(k\) filtering and is equivalent to using the full vocabulary.}
    \begin{tabular}{lcc}
        \toprule
        \textbf{Hyper-parameters} & \shortstack{\textbf{RL of}\\\textbf{QwenVL Model Series}} & \shortstack{\textbf{RL of}\\\textbf{InternVL Model Series}} \\
        \midrule
        \multicolumn{3}{c}{ \textbf{Training Parameters}} \\
        \hline
        epochs & 5 & 5 \\
        batch size & 256 & 256 \\
        clip\_ratio $\epsilon$ & 0.2 & 0.2 \\
        reward weight $\alpha$ & 0.5 & 0.5 \\
        KL weight $\beta$ & 0.005 & 0.0 \\
        Entropy weight $\gamma$ & 0.0 & 0.01 \\
        LR & 3e-6 & 3e-6 \\
        LR warmup steps & 30 & 30 \\
        LR scheduler name & cosine & cosine \\
        max\_response\_length & 1024 & 1024\\
        n\_samples & 16 & 16 \\
        per device train batch size & 32 & 32\\
        dtype & float32 & float32 \\
        \hline
        \multicolumn{3}{c}{\textbf{Rollout Parameters}} \\
        \hline
        temperature & 0.7 & 0.7 \\
        top\_p & 0.9 & 0.7 \\
        top\_k & 50 & -1 \\
        context size & 32768 & 32768 \\
        per device rollout batch size & 32 & 32\\
        dtype & float16 & float16 \\
        \bottomrule
    \end{tabular}
    
    \label{tab:RLpara}
\end{table}
We report the hyperparameters for RL training and rollout in Table~\ref{tab:RLpara}.

\begin{table}[ht]
\setlength{\tabcolsep}{15pt}
\centering
\caption{\textbf{RL vanilla grounding training dataset sizes}. 
All datasets are constructed using the learnability-based selection method detailed in Section~\ref{sec:rl_data_curation}. 
Dataset sizes vary across models due to differences in learnability distributions. }
\begin{tabular}{lc}
\toprule
Model & \#Samples \\
\midrule
InternVL-3-1B   & 100{,}600 \\
InternVL-3-2B   & 63{,}887 \\
InternVL-3-8B   & 37{,}574 \\
\midrule
Qwen3-VL-2B-Thinking   & 111{,}093 \\
Qwen3-VL-4B-Thinking   & 79{,}319 \\
Qwen3-VL-8B-Thinking   & 83{,}989 \\
\bottomrule
\end{tabular}

\label{tab:RLdata_num}
\end{table}
The RL vanilla grounding training dataset sizes are presented in Table~\ref{tab:RLdata_num}. 
The dataset is constructed strictly following the learnability-based procedure described in Section~\ref{sec:rl_data_curation}.

\subsection{Details of Model Inference }
\label{app:detaileval}

\begin{table}[ht]
    \centering
    \caption{\textbf{Parameters of model inference.} }
    \begin{tabular}{lc}
        \toprule
        \textbf{Parameters} & \ \textbf{Inference} \\
        \midrule
        temperature & 0.0 \\
        top\_p & 1.0 \\
        top\_k & 20 \\
        max length & 4096 \\
        repetition penalty & 1.0 \\
        \bottomrule
    \end{tabular}
    
    \label{tab:modelevalution}
\end{table}
We report the inference parameters in Table~\ref{tab:modelevalution}. 
These settings are used consistently throughout the RL training data construction process described in Section~\ref{sec:rl_data_curation}, as well as for all evaluation procedures in Section~\ref{sec:experiments}. 
To ensure fairness, we apply the same inference configuration to all models, which may lead to minor deviations from the official reported results for certain models.

\begin{figure}[ht]
	\centering
\includegraphics[width=1.0\linewidth]{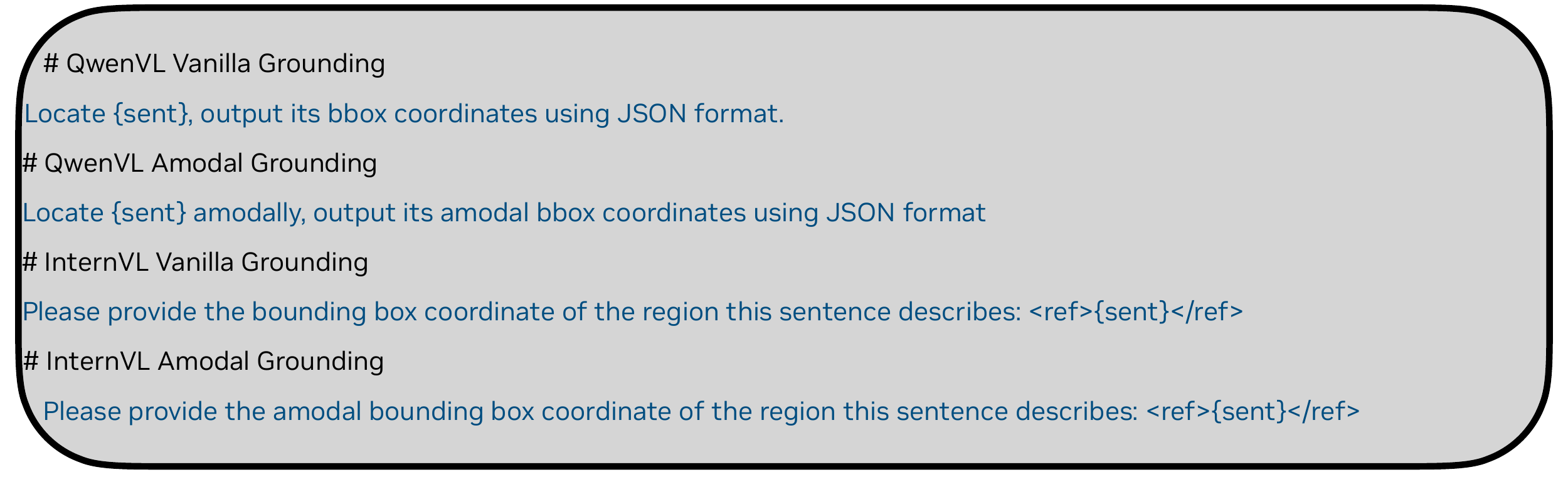}
\caption{\textbf{Prompt for training and inference of models from two different families}.  }
\label{fig:promptfortask}
\end{figure}
In Figure~\ref{fig:promptfortask}, we present the prompts used for two models across the two tasks. 
All vanilla grounding prompts are directly from the reference examples provided in the official GitHub repository, 
while the amodal grounding prompts are derived by modifying the corresponding vanilla grounding prompts.

\clearpage
\section{Details of Grounding Data Curation}
\label{app:detailsftdata}
As mentioned in Section~\ref{sec:method} of the main paper, in this section we provide details of the data curation for vanilla and amodal training data.

\begin{figure}[ht]
	\centering
\includegraphics[width=1.0\linewidth]{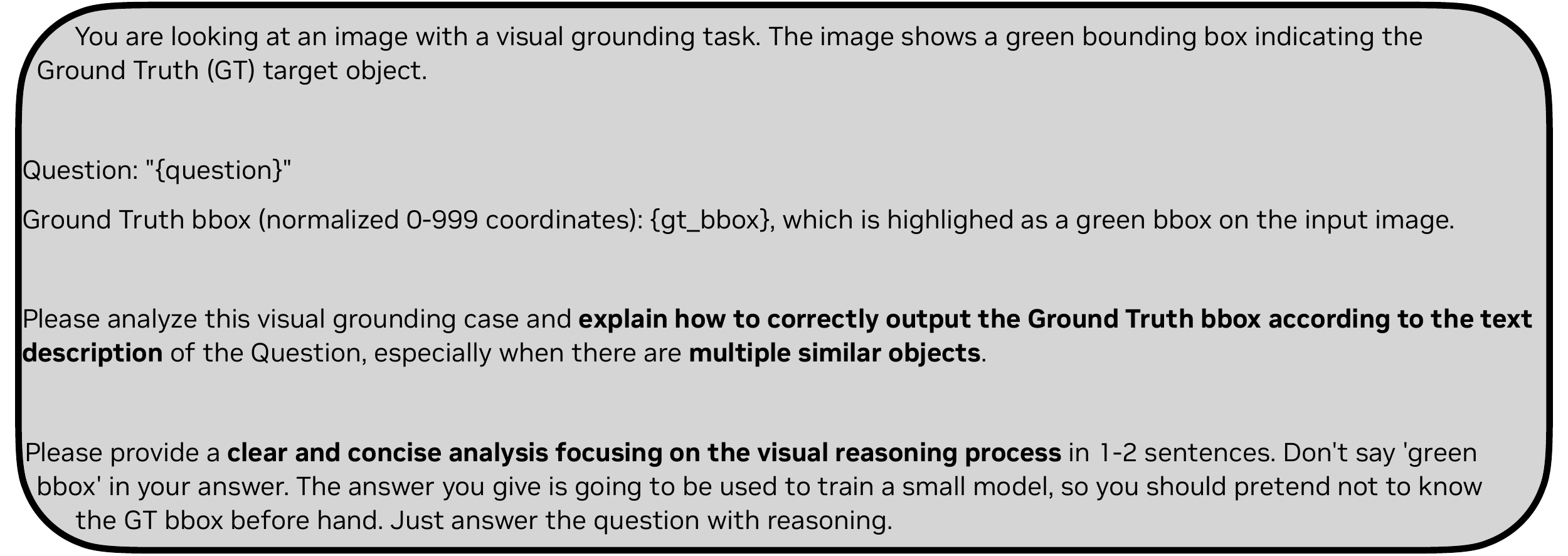}
\caption{\textbf{Prompt for vanilla grounding reasoning dataset generation}.}
\label{fig:supple_prompt_vanilla_reasoning}
\end{figure}

\paragraph{Details of Vanilla Grounding Data Curation.} As in Figure~\ref{fig:supple_prompt_vanilla_reasoning}, we provide the prompt to the proprietary VLM for the generation of the reasoning data of vanilla grounding training.

\begin{figure}[ht]
	\centering
\includegraphics[width=1.0\linewidth]{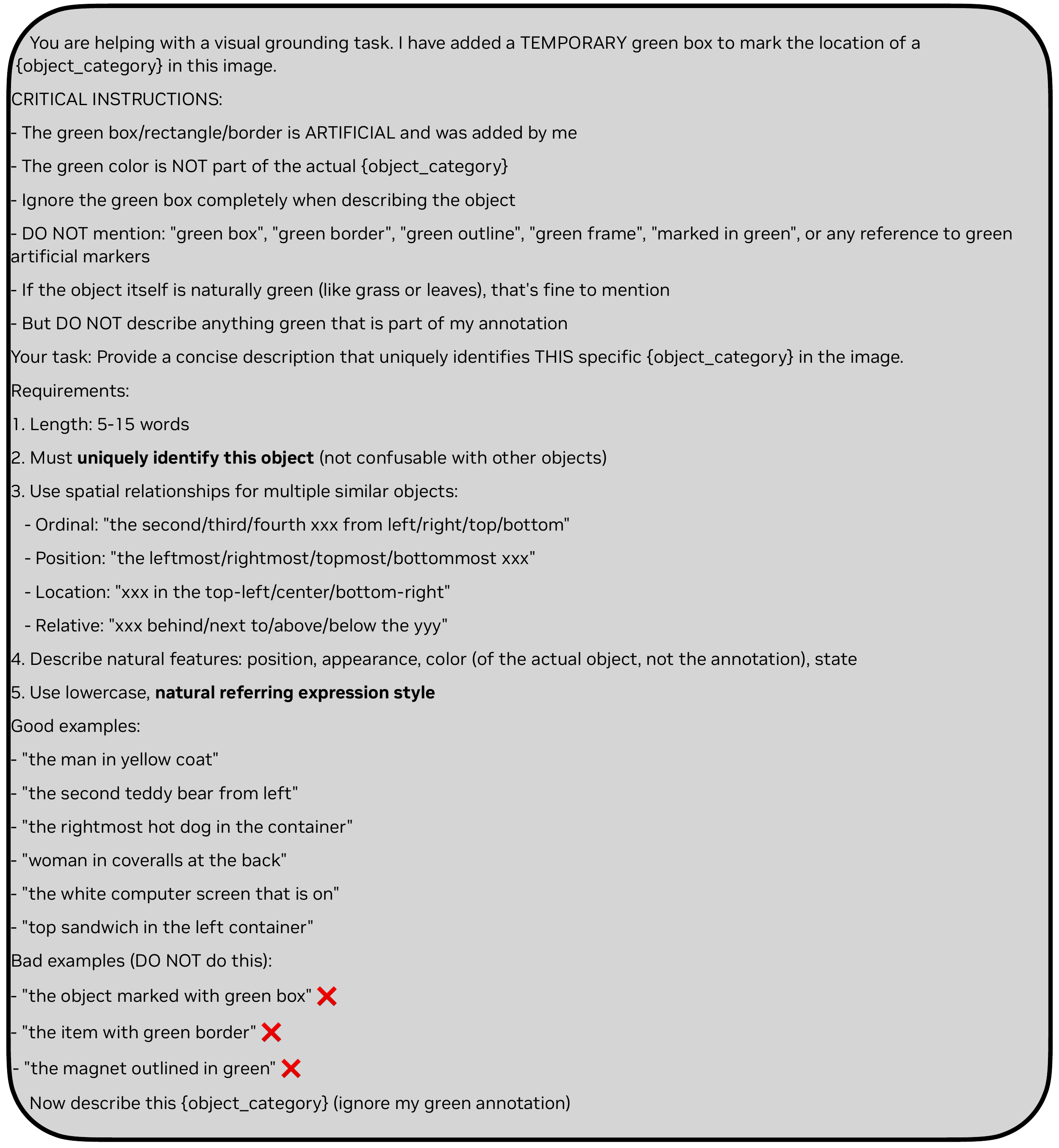}
\caption{\textbf{Prompt for amodal grounding prompt generation}.}
\label{fig:supple_prompt_amodal_prompt_generation}
\end{figure}

\begin{figure}[ht]
	\centering
\includegraphics[width=1.0\linewidth]{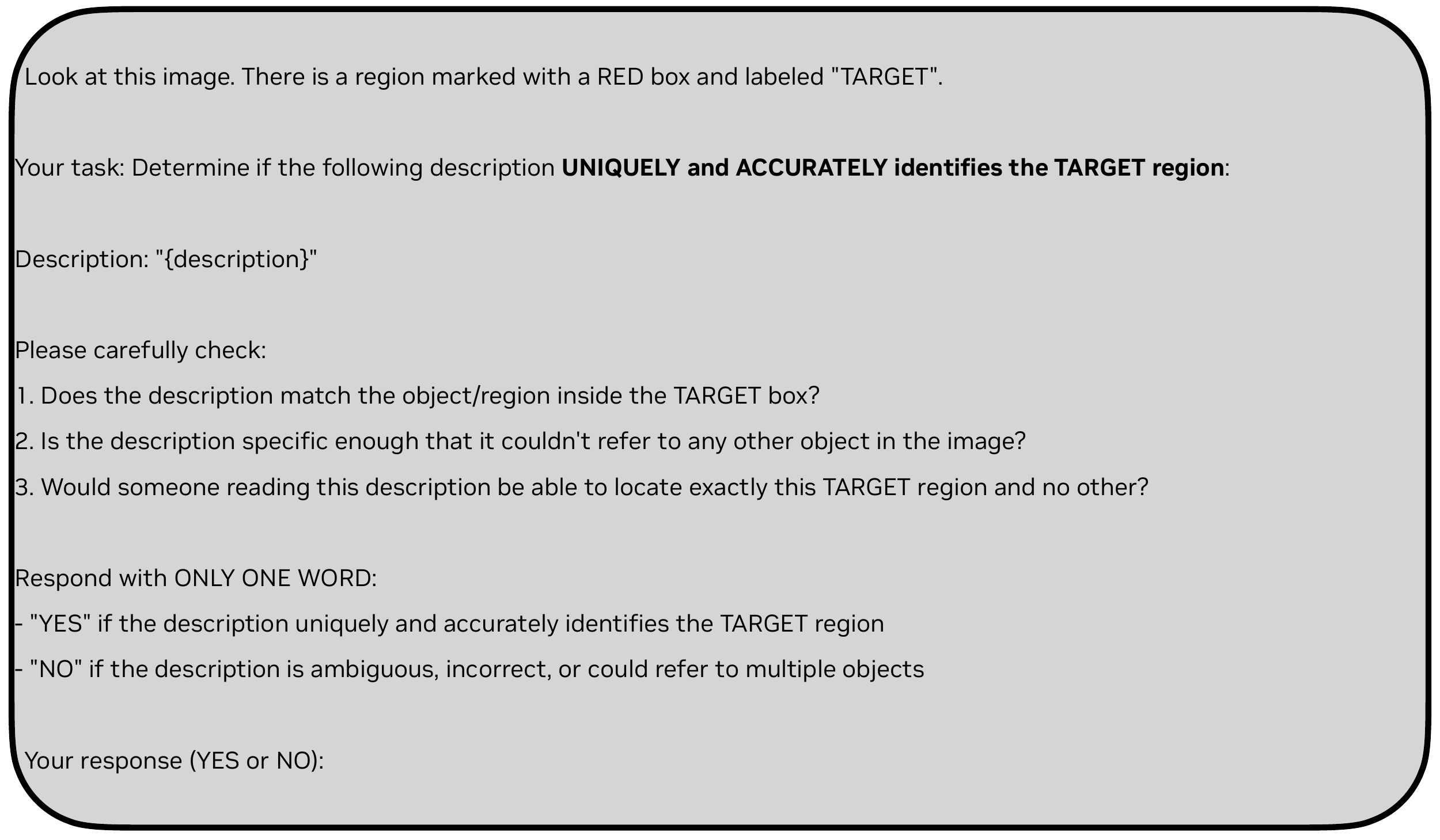}
\caption{\textbf{Prompt for amodal grounding prompt verification}. }
\label{fig:supple_prompt_amodal_prompt_verification}
\end{figure}

\paragraph{Details of Amodal Grounding Data Curation.} 
Different from vanilla grounding, there is no off-the-shelf data for training that provides both the text prompt and the corresponding amodal bounding box. Existing amodal datasets, such as COCO-Amodal~\cite{zhu2017semantic} and TAO-Amodal~\cite{hsieh2023tracking}, provide ground truth annotations of amodal bounding box $B_i$ of occluded objects, but there is only a short category name $C_i$ for each object and no text prompt that can uniquely specify the target object in the image. To bridge the gap, we ask a proprietary VLM, to generate a more detailed description $T_i$ of the target object in the RefCOCO prompt style. Figure~\ref{fig:supple_prompt_amodal_prompt_generation} displays the prompt to the proprietary VLM for text prompt generation.
The generated text prompt can be noisy and we further conduct a verification step: we feed the generated text prompt $T_i$, object bounding box $B_i$ and the image $I_i$ into the proprietary VLM, and ask the VLM to verify whether the generated text prompt can accurately and uniquely specify the target object in the image. This verification step filtered out about $50\%$ of the generated samples. The prompt for the verification step is in Figure~\ref{fig:supple_prompt_amodal_prompt_verification}.

\begin{figure}[ht]
	\centering
\includegraphics[width=1.0\linewidth]{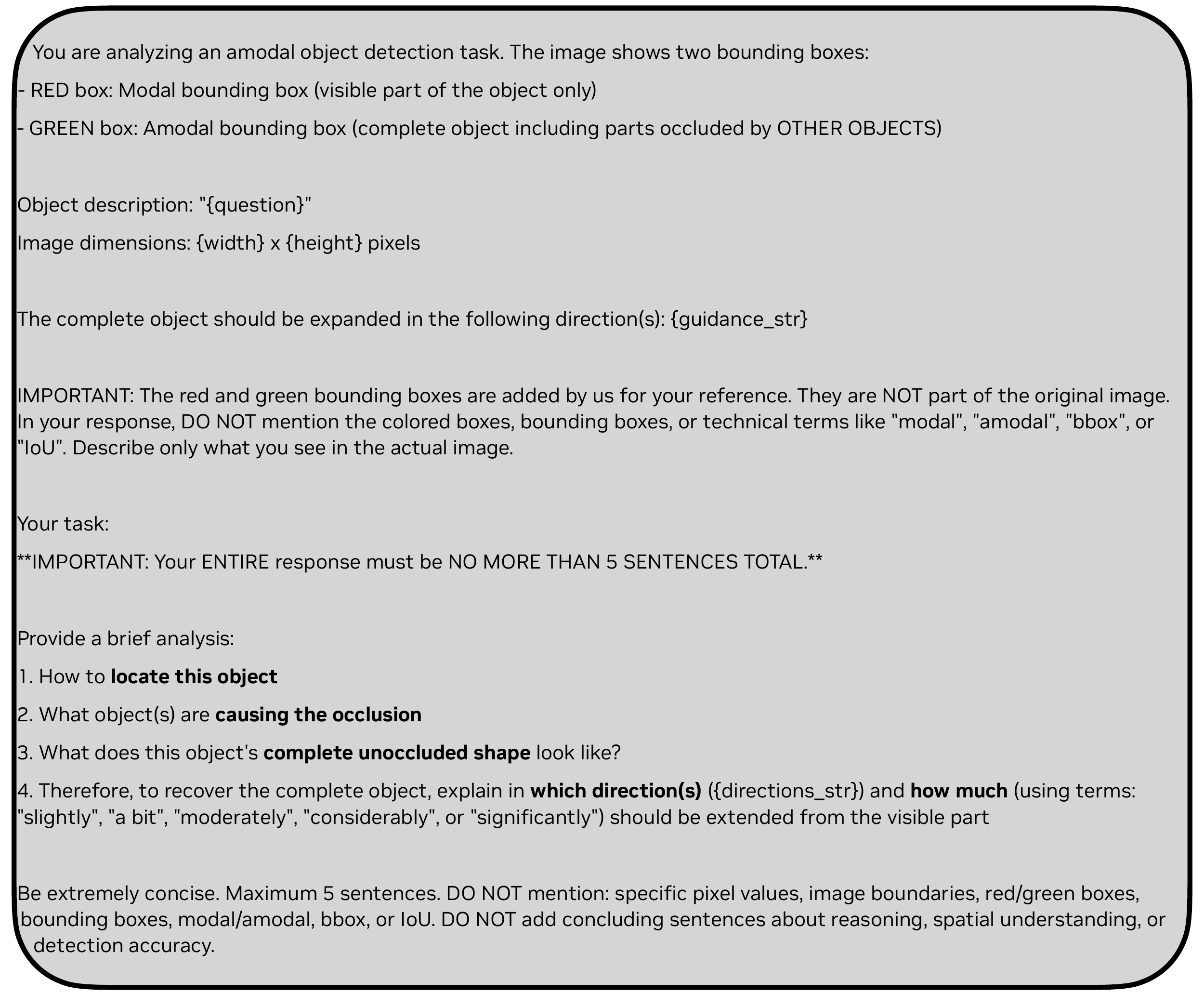}
\caption{\textbf{Prompt for amodal grounding reasoning dataset generation}.  }
\label{fig:supple_prompt_amodal_reasoning}
\end{figure}

Once we obtain the text prompt $T_i$ associated with amodal bounding box $B_i$, we generate the reasoning process $R_i$ via the same pipeline as introduced for vanilla grounding. The only difference is, in addition to how to locate the target object, the reasoning should also involve 1) what object(s) are causing the occlusion; 2) what this object's complete unoccluded shape looks like; 3) therefore, to recover the complete object, in which direction(s) and how much should be extended from the visible part. The detailed prompt to generate the amodal reasoning process is in Figure~\ref{fig:supple_prompt_amodal_reasoning}.

\clearpage

\section{Qualitative Results}
\label{app:qualitative_results}

\begin{figure*}[t]
	\centering
\includegraphics[width=1.0\linewidth]{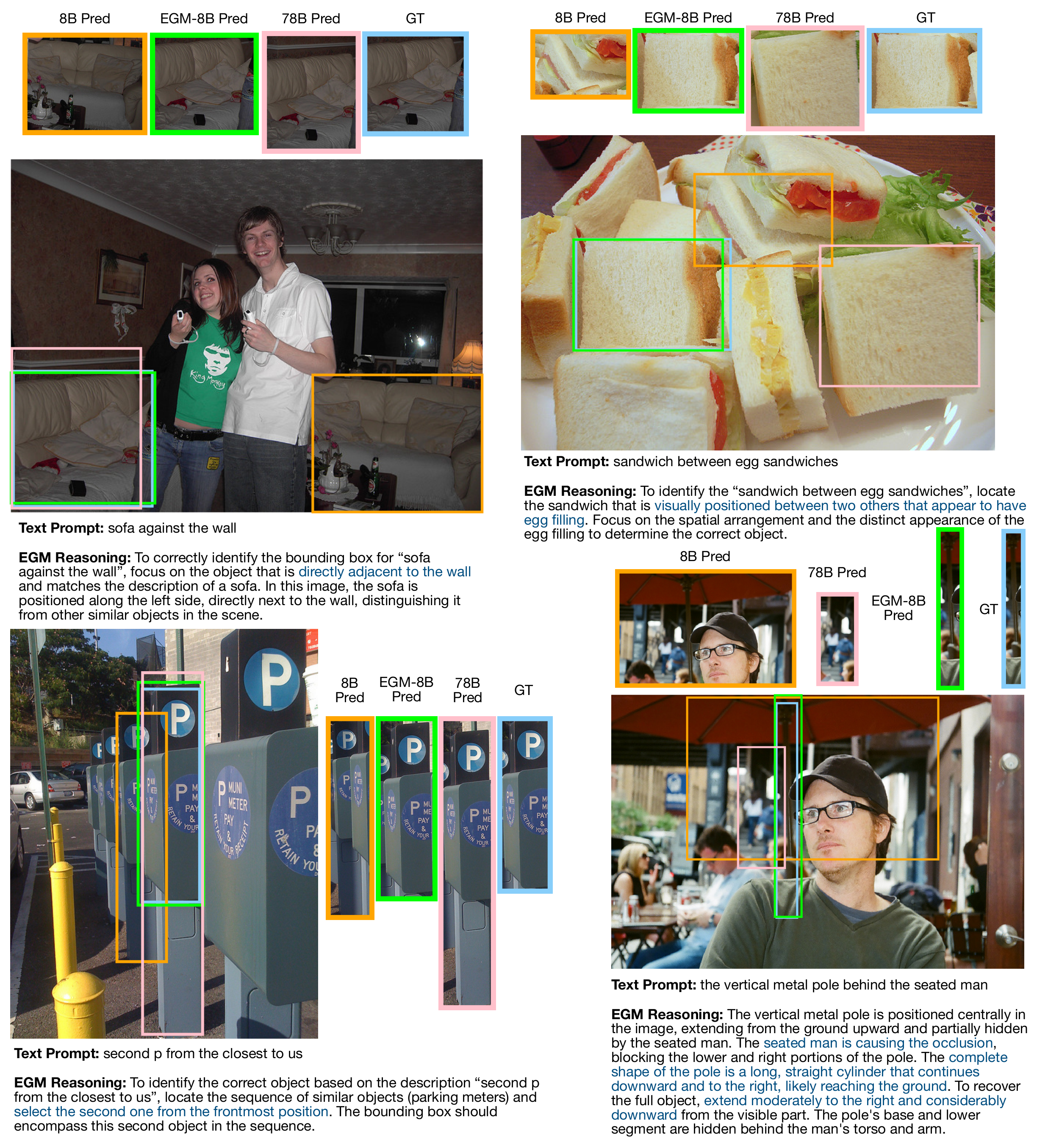}
\caption{
\textbf{Qualitative comparison} of InternVL-3-8B, InternVL-3-78B, and our \method-InternVL-3-8B for vanilla grounding and amodal grounding. Vanilla grounding: top left, top right and bottom left; Amodal grounding: bottom right.
} 
\label{fig:qualitative}
\end{figure*}

As mentioned in Section~\ref{sec:exp_qualitative} of the main paper, in this section we show more qualitative results of our method on the Qwen family (Figure~\ref{fig:qualitative_qwen3}) and InternVL family (Figure~\ref{fig:qualitative}).

Figure~\ref{fig:qualitative_qwen3} shows the qualitative comparison of Qwen3-VL-8B, Qwen3-VL-235B, and our \method-Qwen3-VL-8B for vanilla grounding and amodal grounding. Besides the analysis in the main paper Section~\ref{sec:exp_qualitative}, here we also detail the reasoning process of the Qwen3-VL-235B model.

The Qwen3-VL-235B model also has its reasoning for the thinking model, but the reasoning process is much less efficient than ours. For the top left example of Figure~\ref{fig:qualitative_qwen3}, the entire reasoning of the Qwen3 model is ``Got it, let's see. The problem is to find the girl third from the right in the front row. First, I need to identify the front row. The front row is the group of girls standing in the front, not the ones behind. Let's count from the right. The rightmost person is the one standing furthest to the right. Let's list them: first, second, third. Let's list the front row girls: Looking at the image, the front row (the main group) has several girls. Let's count from the right: 1. Rightmost: girl with white tank top, green skirt, holding racket. 2. Second from right: girl with white cap, green skirt, holding racket. 3. Third from right: the one in white tank top, green skirt, between the cap girl and the next. Wait, let's check. Wait, the front row's count differs. Let's see: front row left to right in front row: 1. Girl in green shirt and white skirt (coach maybe?). 2. Then the girls in white tops and green skirts. Let's count the players in the front row (excluding the coach). Let's see: from right to left: 1. Rightmost player: girl with white top, green skirt, holding racket. 2. Second: girl with white cap, green skirt, holding racket. 3. Third from right: the girl in white top, green skirt, between the cap girl and the next. Let's check the image. Looking at the image, let's count from the right: - First (rightmost): [838, 129, 993, 999] - Second: [742, 299, 922, 999] - Third from right: [635, 711, 771, 999]. Yes, that's the `third from the right'. Let's confirm. The third from the right in the front row (players) is the girl with the white top, green skirt, and she's standing next to the girl with the white cap. So the bounding box for that girl is [635, 711, 771, 999].'' It can be observed that the reasoning process contains lots of redundant words, and therefore is much less efficient than our \method~reasoning, which concisely focuses on the key feature that distinguishes the target object, as displayed on the bottom of the example in Figure~\ref{fig:qualitative_qwen3} top left.

\clearpage

Figure~\ref{fig:qualitative} shows the qualitative comparison of InternVL-3-8B, InternVL-3-78B, and our \method-InternVL-3-8B for vanilla grounding and amodal grounding. 

The top left, top right and bottom left samples are vanilla grounding on RefCOCO benchmark. For the \emph{top left} example, the text prompt is `sofa against the wall', and the InternVL-3-8B model mis-locates to another sofa. Our \method-InternVL-3-8B correctly reasons the key feature `directly adjacent to the wall' and generates a correct prediction. For the \emph{top right} example, the text prompt is `sandwich between egg sandwiches', and the InternVL-3-8B model confuses it with another sandwich. Our \method-InternVL-3-8B, however, correctly reasons about the key feature `visually positioned between two others that appear to have egg filling' and outputs the correct grounding box. For the \emph{bottom left} example, the text prompt is `second P from the closest to us', and the InternVL-3-8B model fails to understand `second' and mis-predicts the `third' P. Our \method-InternVL-3-8B correctly predicts the correct `second P', with a reasoning that it should be the `second one from the frontmost position'. For the vanilla grounding cases, generally InternVL-3-78B performs better than InternVL-3-8B, as can be observed in the examples.

The \emph{bottom right} example is amodal grounding on the COCO-Amodal benchmark. It can be observed that InternVL-3-8B and InternVL-3-78B fail to predict the amodal grounding box of `the vertical metal pole behind the seated man', while our \method-InternVL-3-8B prediction is close to the ground truth, with reasoning process about locating the occluded metal pole, analyzing it is occluded by the seated man and should extend downward to recover the full unoccluded shape.
This reasoning helps \method-InternVL-3-8B to precisely locate the amodal bounding box covering both the visible and occluded parts of the metal pole.

\clearpage
\section{Prompt for Failure Case Analysis}
\label{app:prompt_failure_case_analysis}

\begin{figure}[h]
	\centering
\includegraphics[width=1.0\linewidth]{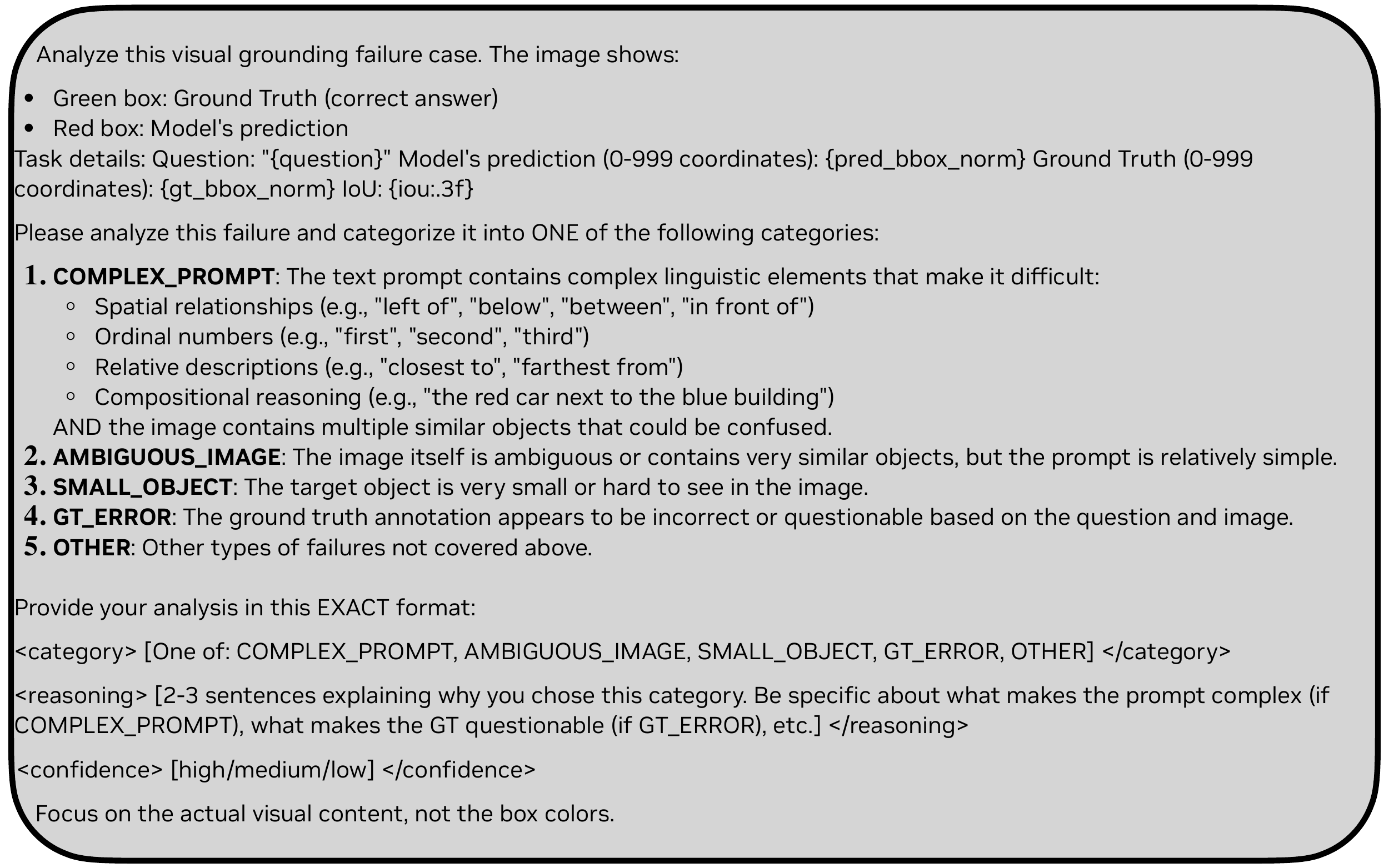}
\caption{
\textbf{The prompt to analyze failure reasons of small VLMs.}
} 
\label{fig:failure_case_analysis_gpt_prompt}
\end{figure}

As mentioned in Section~\ref{sec:problem_analysis} of the main paper, Figure~\ref{fig:failure_case_analysis_gpt_prompt} shows the prompt we used to analyze the failure reason of small VLMs. It can be seen that the failure reasons are categorized into 5 categories: `COMPLEX-PROMPT', `AMBIGUOUS-IMAGE', `SMALL-OBJECT', `GT-ERROR' and `Other'.

\clearpage
\section{Ablation Study of SFT and RL}
\label{app:ablation_sft_rf}

As mentioned in Section~\ref{sec:exp_vanilla_grounding} of the main paper, Table~\ref{tab:ablation_sft_rl_gains} shows the ablation results of SFT training and RL training on Qwen3-VL models. It can be observed that the SFT stage brings improvement of the models, and the RL stage brings further improvement.

\begin{table}[h]
    \centering
    \small
    \renewcommand{\arraystretch}{1.2}
    \setlength{\tabcolsep}{10pt}
    \caption{\textbf{Ablation study of SFT and RL improvements.} The green subscripts denote the improvement compared to the previous stage.}
    \label{tab:ablation_sft_rl_gains}
    \begin{tabular}{l c c c}
        \toprule
        \textbf{Model} & \textbf{Base} & \textbf{+ SFT} & \textbf{+ RL} \\
        \midrule
        2B & 83.6 & $85.6_{\color{green!60!black}\uparrow 2.0}$ & $\textbf{89.6}_{\color{green!60!black}\uparrow 4.0}$ \\
        4B & 87.2 & $87.9_{\color{green!60!black}\uparrow 0.7}$ & $\textbf{91.0}_{\color{green!60!black}\uparrow 3.1}$ \\
        8B & 87.8 & $88.8_{\color{green!60!black}\uparrow 1.0}$ & $\textbf{91.4}_{\color{green!60!black}\uparrow 2.6}$ \\
        \bottomrule
    \end{tabular}
\end{table}

\clearpage
\section{Ablation Study of Reasoning Process}
\label{app:ablation_reasoning_process}

As a supplement to Section~\ref{sec:exp_vanilla_grounding}, we provide the full ablation results for the reasoning process in Table~\ref{tab:ablation_reasoning}. To remove the model’s ability to generate reasoning process, we enforce a JSON schema that restricts intermediate reasoning outputs. The results show that reasoning process consistently improves performance across models of different sizes and families. In particular, the changes observed for InternVL-3-8B ($-1.9/+1.1$) and Qwen3-VL-8B-Thinking ($+0.8/+3.6$) suggest that the performance gains from the reasoning process tend to be more significant for larger models.

\begin{table*}[h]
    \centering
    \setlength{\tabcolsep}{5pt}
    \renewcommand{\arraystretch}{1.2}
    \caption{\textbf{Ablation results of the reasoning process. } The Comparison of our EGM method with and without the reasoning process (\textit{w/o R}) indicates that the reasoning process significantly enhances performance of models of different sizes and from different families.
    Entries marked with * are from the official report~\cite{wang2025internvl35}, and the others are measured by ourselves.
    }

    \resizebox{\linewidth}{!}{
    \begin{tabular}{l c c c c c c c c l}
    \toprule[1.5pt]
    \multirow{2}{*}{Model} & \multicolumn{3}{c}{RefCOCO} & \multicolumn{3}{c}{RefCOCO+} & \multicolumn{2}{c}{RefCOCOg} & \multirow{2}{*}{Avg. Acc} \\
     & val & test-A & test-B & val & test-A & test-B & val & test &  \\
     \midrule[1.2pt]
      Qwen3-VL-2B-Thinking & 87.5 & 91.4 & 82.0 & 80.2 & 86.7 & 72.1 & 84.4 & 84.5 & 83.6 \\
      \rowcolor{black!10} \quad + \textit{\method w/o R}  & 92.2 & 93.5 & 89.0 & 86.3 & 90.9 & 80.8 & 88.6 & 89.0 & ${\text{88.7}}_{\color{yellowgreen}+5.1\uparrow}$ \\
      \rowcolor{black!10} \quad + \textit{\method}  & 93.0 & 94.0 & 89.4 & 87.8 & 91.7 & 82.7 & 88.6 & 89.3 & ${\text{89.6}}_{\color{yellowgreen}+6.0\uparrow}$ \\
    \cmidrule(lr){1-10}
      Qwen3-VL-4B-Thinking & 90.0 & 92.7 & 85.6 & 85.2 & 89.5 & 79.3 & 87.8 & 87.7 & 87.2 \\
      \rowcolor{black!10} \quad + \textit{\method w/o R}  & 93.5 & 95.0 & 90.8 & 89.3 & 92.4 & 83.8 & 90.1 & 90.7 & ${\text{90.7}}_{\color{yellowgreen}+3.5\uparrow}$ \\
      \rowcolor{black!10} \quad + \textit{\method}  & 93.5 & 95.1 & 90.9 & 89.7 & 93.1 & 84.9 & 90.4 & 90.8 & ${\text{91.0}}_{\color{yellowgreen}+3.8\uparrow}$ \\
    \cmidrule(lr){1-10}
      Qwen3-VL-8B-Thinking & 91.0 & 92.9 & 86.9 & 86.2 & 89.3 & 80.2 & 87.6 & 88.6 & 87.8 \\
      \rowcolor{black!10} \quad + \textit{\method w/o R}  & 92.2 & 93.6 & 89.2 & 85.7 & 90.2 & 80.0 & 88.6 & 89.3 & ${\text{88.6}}_{\color{yellowgreen}+0.8\uparrow}$ \\
      \rowcolor{black!10} \quad + \textit{\method}  & 93.9 & 95.0 & 91.2 & 90.1 & 93.3 & 85.9 & 90.4 & 91.2 & ${\textbf{\text{91.4}}}_{\color{yellowgreen}+3.6\uparrow}$ \\
    \midrule[1.2pt]
    InternVL-3-1B & 85.8 & 90.1  & 81.7  & 76.6  & 84.1  & 69.2 & 82.8  & 82.6  & 81.6* \\
    \rowcolor{black!10} \quad + \textit{\method w/o R}  & 87.7 & 91.5 & 83.9 & 79.7 & 86.3 & 72.6 & 83.6 & 84.1 & ${\text{83.6}}_{\color{yellowgreen}+2.0\uparrow}$ \\
      \rowcolor{black!10} \quad + \textit{\method}  & 90.2  & 93.2 & 87.0 & 83.8 & 88.8 & 77.5 & 86.4 & 87.5 & ${\text{86.8}}_{\color{yellowgreen}+5.2\uparrow}$ \\
    \cmidrule(lr){1-10}
      InternVL-3-2B & 89.8 & 92.6  & 86.4  & 84.0  & 89.2 & 76.5  & 87.6  & 87.2 & 86.7* \\
      \rowcolor{black!10} \quad + \textit{\method w/o R}  & 90.6 & 93.1 & 86.7 & 84.6 & 89.9 & 77.5 & 86.7 & 87.2 & ${\text{87.0}}_{\color{yellowgreen}+0.3\uparrow}$ \\
      \rowcolor{black!10} \quad + \textit{\method}  & 92.2 & 94.0 & 87.4 & 85.6 & 91.2 & 79.2 & 88.5 & 88.7 & ${\text{88.4}}_{\color{yellowgreen}+1.7\uparrow}$ \\
    \cmidrule(lr){1-10}
      InternVL-3-8B & 92.5 & 94.6  & 88.0  & 88.2  & 92.5 & 81.8  & 89.6 & 90.0 & 89.6* \\
      \rowcolor{black!10} \quad + \textit{\method w/o R}  & 91.3 & 93.9 & 87.0 & 85.7 & 90.6 & 78.1 & 87.3 & 88.2 & ${\text{87.7}}_{\color{red}-1.9 \downarrow}$ \\
      \rowcolor{black!10} \quad + \textit{\method}  & 93.6 & 95.2 & 90.1 & 89.3 & 93.6 & 83.1 & 89.7 & 90.7 & ${\text{90.7}}_{\color{yellowgreen}+1.1\uparrow}$ \\
    \bottomrule[1.5pt]
    \end{tabular}
    }

    \label{tab:ablation_reasoning}
    \end{table*}

\clearpage
\section{Performance Gains on Hard Samples}
\label{app:hard_sample}

Beyond the analysis in Section~\ref{sec:exp_vanilla_grounding}, we conduct a fine-grained analysis of performance gains on difficult instances from the RefCOCO benchmark~\cite{kazemzadeh2014referitgame,mao2016generation} using Qwen3-VL-8B-Thinking as the base model. We define hard samples as those with baseline IoU$<0.5$. After applying our method, we re-evaluate these samples and categorize them into two bins according to the resulting IoU: \emph{Unsolved} (IoU$\in[0,0.5)$) and \emph{Solved} (IoU$\in[0.5,1.0]$).

Table~\ref{tab:hard_sample_binary_flow} shows that our method resolves a substantial portion of previously hard cases. Specifically, among 6,082 samples with baseline IoU$<0.5$, 3,030 (49.8\%) are upgraded to IoU$\ge 0.5$, while 3,052 (50.2\%) remain below 0.5. These results suggest that the proposed approach markedly improves grounding accuracy on challenging instances.

\begin{table}[h]
\centering
\caption{\textbf{Hard-sample flow (baseline IoU$<0.5$, $N{=}6{,}082$).} We count a case as solved if the post-method IoU is at least 0.5.}
\label{tab:hard_sample_binary_flow}
\scalebox{1.15}{
\begin{tabular}{lrr}
\hline
\textbf{Post IoU} & \textbf{Count} & \textbf{\%} \\
\hline
$[0, 0.5)$   & 3,052 & 50.2 \\
$[0.5, 1.0]$ & 3,030 & 49.8 \\
\hline
\end{tabular}}
\end{table}

\clearpage
\section{Analysis of Failure Cases of Small VLMs}
\label{app:detail_analysis_failure_case_small_vlm}

As mentioned in Section~\ref{sec:problem_analysis} of the main paper, in Table~\ref{tab:failure_error_distribution_entire} we provide the entire analysis of failure cases of the InternVL-3-8B model using different proprietary models. It can be observed that a large proportion of the failure cases of the small VLM are due to~\failreason, and this is agreed by different proprietary models.

\begin{table}[htbp]
\centering
\setlength{\tabcolsep}{10pt}
\caption{\textbf{Entire analysis of failure cases of small VLM} using different proprietary models.}
\begin{tabular}{lccc}
\toprule
\textbf{Failure Reason} & \textbf{GPT-4} & \textbf{GPT-5} & \textbf{Gemini-3-Pro} \\
\midrule
COMPLEX-PROMPT & 62.8\% & 48.8\% & 46.8\% \\
AMBIGUOUS-IMAGE & 35.9\% & 21.7\% & 17.9\% \\
SMALL-OBJECT & 0.4\% & 1.0\% & 0.2\% \\
GT-ERROR & 0.6\% & 11.0\% & 27.6\% \\
OTHER & 0.3\% & 17.5\% & 7.5\% \\
\bottomrule
\end{tabular}

\label{tab:failure_error_distribution_entire}
\end{table}